\documentclass[10pt,twocolumn,letterpaper]{article}

\usepackage[pagenumbers]{cvpr} 

\definecolor{cvprblue}{rgb}{0.21,0.49,0.74}
\usepackage[pagebackref,breaklinks,colorlinks,allcolors=cvprblue]{hyperref}
\usepackage{soul}
\def\paperID{39850} 
\def\confName{CVPR}
\def\confYear{2026}

\title{Bright 4B: Scaling Hyperspherical Learning for Segmentation\\ in 3D Brightfield Microscopy}

\author{
  Amil Khan\textsuperscript{1} \and
  Matheus Viana\textsuperscript{2} \and
  Suraj Mishra\textsuperscript{2} \and
  B.S. Manjunath\textsuperscript{1} \\[1.5ex]
  \textsuperscript{1}UC Santa Barbara \quad
  \textsuperscript{2}Allen Institute for Cell Sciences \\
  {\tt\small \{amil, manj\}@ucsb.edu, \{matheus.viana, suraj.mishra\}@alleninstitute.org}
}
\usepackage[svgnames]{xcolor}
\usepackage{tcolorbox}
\newcommand{\ws}[1]{%
  \colorbox{LightGreen}{#1}%
}
\newcommand{\rd}[1]{%
  \colorbox{MistyRose}{#1}%
}

\newcommand{\sm}[1]{{\color{MediumSeaGreen}{SM: #1}}}
\begin{document}
\twocolumn[{%
\renewcommand\twocolumn[1][]{#1}%
\maketitle
\begin{center}
    \centering
    \captionsetup{type=figure}
    \includegraphics[width=1\textwidth]{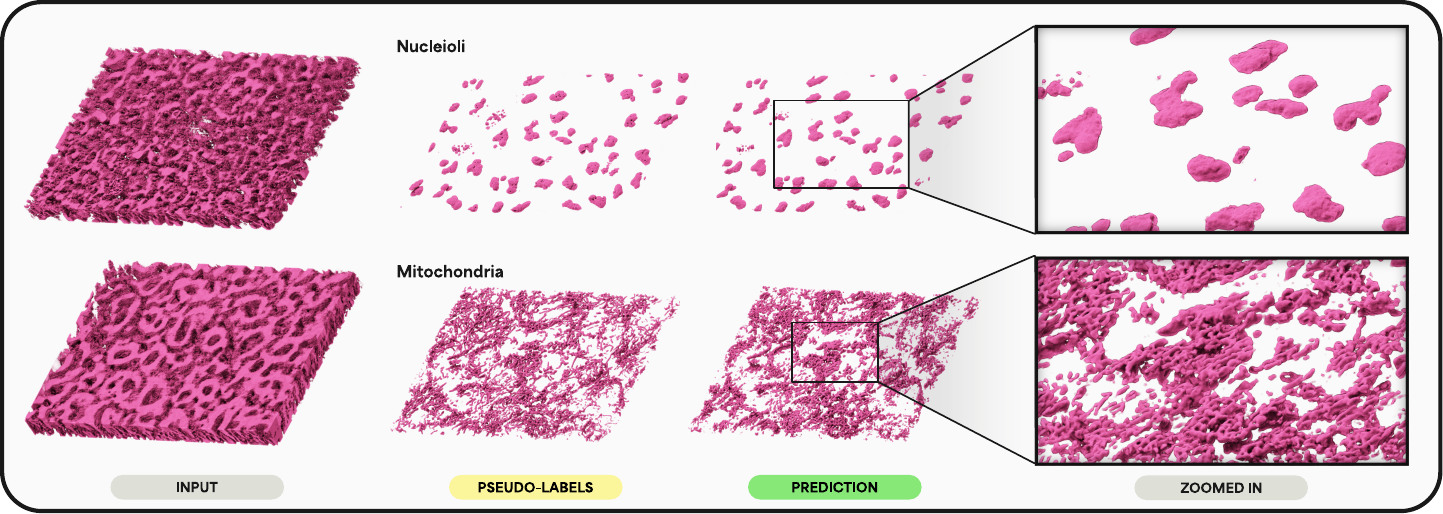}
    \captionof{figure}{\textbf{Bright 4B} is a 4B-parameter foundation model for segmenting subcellular structures in 3D brightfield microscopy images that learns on the unit hypersphere. It accepts label-free 3D brightfield microscopy images as input and predicts subcellular structures in 3D, preserving fine structural detail across depth and cell types, without requiring explicit fluorescence images.}
\end{center}%
}]

\begin{abstract}



Label-free 3D brightfield microscopy offers a fast and noninvasive way to visualize cellular morphology, yet robust volumetric segmentation still typically depends on fluorescence or heavy post-processing. We address this gap by introducing \textbf{Bright-4B}---a 4B-parameter foundation model that learns on the unit hypersphere to segment subcellular structures directly from 3D brightfield volumes. Bright-4B combines a hardware-aligned \emph{Native Sparse Attention} mechanism (capturing local, coarse, and selected global context), depth--width residual \emph{HyperConnections} that stabilize representation flow, and a soft \emph{Mixture-of-Experts} for adaptive capacity. A plug-and-play \emph{anisotropic patch embed} further respects confocal point-spread and axial thinning, enabling geometry-faithful 3D tokenization. The resulting model produces morphology-accurate segmentations of nuclei, mitochondria, and other organelles from brightfield stacks alone---without fluorescence, auxiliary channels, or handcrafted post-processing. Across multiple confocal datasets, Bright-4B preserves fine structural detail across depth and cell types, outperforming contemporary CNN and Transformer baselines. All code, pretrained weights, and models for downstream finetuning will be released to advance large-scale, label-free 3D cell mapping.

\end{abstract}    
\section{Introduction}
\label{sec:intro}

\begin{figure}
    \centering
    \includegraphics[width=1\linewidth]{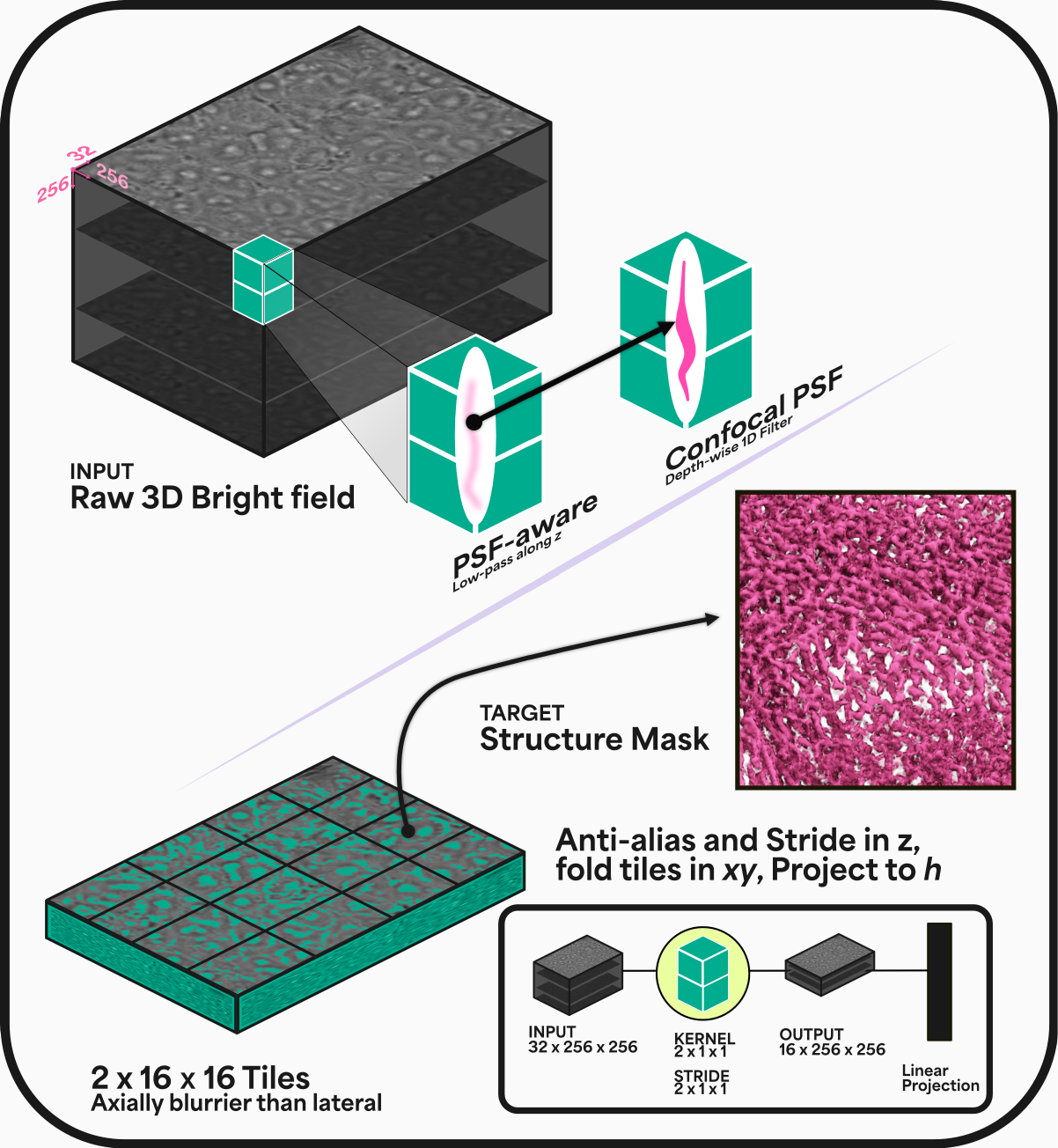}
    \caption{\textbf{Anisotropic 3D patch embedding.} A PSF-aware axial low-pass with stride $p_z$ prevents aliasing, followed by tiling in the $xy$ plane ($16\times16$). A bias-free projection maps each tile to $h$ dimensions, producing a geometry-faithful token lattice $(B,D',H',W',h)$ with $D'=D/2$. This $(2\times16\times16)$ design respects confocal anisotropy and supports accurate 3D segmentation.}
    \label{fig:placeholder}
\end{figure}

Large-scale vision models have transformed 2D perception, yet native 3D architectures with strong pretrained weights remain limited. Most existing designs still treat volumes as stacks of 2D slices, weakening long-range coherence—especially in microscopy, where axial sampling is anisotropic and brightfield contrast changes sharply with depth.

To address this gap, we revisit transformer design for volumetric data flow~\cite{elhage2021mathematical}. We begin with an \emph{anisotropic 3D patch embed} that applies a PSF-aware low-pass only along $z$ while preserving native in-plane resolution. The resulting $(D' \!\times\!H\!\times\!W)$ token lattice mitigates axial aliasing. Within each block, we adopt \emph{Native Sparse Attention (NSA)}~\cite{yuan2025native}, which mixes (i) local sliding windows capturing brightfield edges, (ii) a coarse compressive pathway aggregating mesoscale context, and (iii) selected global regions. A learned per-head gate blends these routes efficiently.

\textbf{Objective.} We aim to \emph{directly} segment subcellular structures (e.g., nucleoli, mitochondria, Golgi) from raw 3D brightfield volumes. Since brightfield signals exhibit low contrast and depth-dependent artifacts, we design for stable representation learning. All token states and projections are unit-normalized so updates lie on the \emph{unit hypersphere}, improving numerical stability at the 4B-parameter scale. Attention and MoE blocks are wrapped with \emph{Dynamic HyperConnections}, which softly mix residual streams. A load-balanced, slot-based \emph{Soft MoE} enables expert specialization (boundary sharpening, noise suppression, morphology), and a lightweight UNETR-style decoder restores full-resolution masks.

\paragraph{Why brightfield?}
Brightfield microscopy is fast, gentle, and widely available, but accurate 3D segmentation typically still requires fluorescence, extra channels, or handcrafted pipelines. Prior work either relies on fluorescence supervision~\cite{kraus2024masked} or multi-stage heuristics~\cite{stringer2021cellpose,pachitariu2025cellpose,hatamizadeh2022unetr}. Label-free methods mainly predict \emph{fluorescent images} from brightfield~\cite{viana2023integrated}, requiring separate segmentation. Meanwhile, modern vision transformers~\cite{dosovitskiy2020image,dehghani2023scaling,zhai2022scaling} scale effectively, but their windowing patterns and isotropic assumptions are mismatched to brightfield’s axial anisotropy, halo artifacts, and low-contrast depth structure. We instead infer organelle masks \emph{directly} from brightfield, avoiding image-to-image translation.

\paragraph{Bright-4B as a foundation model.}
We target a domain-specific \emph{foundation model} for brightfield microscopy: a high-capacity 3D architecture trained on large-scale brightfield volumes, serving as transferable initialization across organelles, cell types, and imaging regimes.

\begin{enumerate}
    \item \textbf{Bright-4B architecture.} A 4B-parameter 3D transformer that learns subcellular structure on the unit hypersphere. It integrates Native Sparse Attention for multi-scale volumetric reasoning, a normalized Soft MoE for adaptive capacity, and Dynamic HyperConnections for stable residual mixing.

    \item \textbf{Direct label-free 3D segmentation.} Bright-4B predicts 3D masks for multiple organelles from a single brightfield stack, without fluorescence channels or post-processing. Performance varies with intrinsic organelle visibility, yet remains robust across depth and morphology.

    \item \textbf{Open foundation for microscopy AI.} We release pretrained weights, training code, and structure-specific models to support community benchmarking and scaling.
\end{enumerate}

Unlike prior normalized MoEs, where scale drift disrupts routing dynamics, \textbf{Bright-4B} performs attention, routing, and residual updates entirely on the unit hypersphere, eliminating gradient explosion and enabling stable sparse expert behavior at the billion-parameter scale.

\section{Related Work}
\label{sec:relatedworks}

\paragraph{Label-free prediction and nuclear segmentation.}
Early work on label-free fluorescence prediction used CNNs to map transmitted-light images to fluorescent channels~\cite{christiansen2018silico,ounkomol2018label,rivenson2019phasestain}. While successful in 2D, performance degrades in 3D brightfield due to axial blur and depth-dependent ambiguity. Instance-segmentation frameworks such as Cellpose~\cite{stringer2021cellpose} incorporate brightfield modes, but rely on strong morphological priors, 2D processing, or post-hoc 3D reconstruction. Our setting is stricter: we predict \emph{3D subcellular masks directly from brightfield volumes}, end-to-end, without fluorescence, priors, or handcrafted stages.

\paragraph{U-Net families in 3D.}
3D U-Net~\cite{cciccek20163d}, V-Net~\cite{milletari2016v}, UNet++~\cite{zhou2018unet++}, nnU-Net/dynUNet~\cite{Isensee_2020}, and UNETR variants~\cite{hatamizadeh2022unetr} remain strong volumetric baselines. They capture local features well but struggle with confocal anisotropy: axial striding and 3D convolutions often smear thin structures or introduce block artifacts along $z$. Long-range context further requires deep encoders or multi-scale fusion. Bright-4B addresses these limitations with (i) an \textbf{anisotropic patch embedding} that anti-aliases before axial decimation and tiles $xy$ into $(2\times16\times16)$ tokens aligned with the PSF, and (ii) \textbf{Native Sparse Attention} that provides efficient global reasoning without deep pyramids.

\paragraph{Vision Transformers for medical imaging.}
Transformer backbones such as ViT and DeiT, adapted through TransUNet~\cite{chen2021transunet}, UNETR~\cite{hatamizadeh2022unetr}, and hybrids, improve long-range reasoning but incur quadratic cost. Swin Transformer~\cite{liu2021swin} and Swin-UNETR~\cite{hatamizadeh2021swin} mitigate this via shifted windows, yet window locality creates \emph{context barriers}: fine details are sharp within windows, but cross-window or cross-slice coherence propagates slowly and is sensitive to stage parameters. Under confocal anisotropy, such windowing tends to over-aggregate axially or under-connect laterally. Our \textbf{Native Sparse Attention (NSA)} replaces windowing with a three-path routing scheme—local sliding, coarse compressive memory, and selected global blocks—that integrates distant context in a single hop while preserving brightfield-specific edges.

\paragraph{Efficient and sparse attention.}
Longformer, BigBird, and deformable attention~\cite{beltagy2020longformer,zaheer2020big,zhu2020deformable} reduce quadratic cost via structured sparsity or learned sampling. \textbf{NSA}~\cite{yuan2025native}, which we extend from 2D sequences to \emph{3D anisotropic token lattices}, combines local windows, \emph{compressive} memory, and \emph{top-$k$} selected global blocks, with per-head mixing. This adapts naturally to confocal geometry: sliding windows capture local gradients, compression summarizes axial context, and selected blocks recover rare but decisive long-range cues (e.g., nucleolar contrast, cross-slice morphology). Bright-4B is, to our knowledge, the first to deploy NSA on 3D anisotropic volumes.

\paragraph{Stabilizing deep Transformers.}
Pre-norm and RMSNorm~\cite{zhang2019root} improve stability, but deep stacks may still drift in scale or misgate residuals. \textbf{HyperConnections}~\cite{Zhu2024HyperConnections} introduce width and depth gates to interpolate pre- and post-update states across multiple residual streams. Normalized-Transformer formulations~\cite{loshchilov2024ngpt} further motivate constraining representations to the unit hypersphere. Bright-4B combines these ideas: all hidden states and projections are $L_2$-normalized so updates act as hyperspherical rotations, while HyperConnections regulate residual mixing—preserving fine contrast essential for brightfield boundary recovery.

\paragraph{Mixture-of-Experts at scale.}
Sparse MoE models such as GShard and Switch~\cite{lepikhin2020gshard,fedus2022switch} expand capacity with conditional computation, but hard routing is brittle under class imbalance and heterogeneous medical data. Soft MoE~\cite{puigcerver2023sparse} improves gradient flow by using smooth dispatch and combination. We extend this to a \textbf{slot-based SoftMoE} operating on the \emph{unit hypersphere}: tokens form dynamic slots, experts specialize in fine- or low-frequency structure, and outputs are re-normalized. This stabilizes routing at the billion-parameter scale and enables selective computation on faint subcellular features without overfitting to brightfield artifacts.

\section{Methods}

\paragraph{Overview and design synergy.}
Bright-4B unifies four complementary components into a coherent 3D segmentation architecture for label-free brightfield data. The \textbf{anisotropic patch embedding} aligns tokenization with the microscope PSF, so axial information is decimated only after PSF-aware anti-alias filtering. These geometry-faithful tokens feed into \textbf{Native Sparse Attention (NSA)}, which routes information through local, coarse, and global paths matched to brightfield spatial statistics. The resulting multi-scale features are processed by a \textbf{slot-based SoftMoE} that operates entirely on the \textit{unit hypersphere}, enabling stable expert specialization without scale drift. Finally, \textbf{Dynamic HyperConnections} blend multiple residual streams via learned width–depth gates, maintaining representation stability at billion-parameter depth. Together, these modules form a hyperspherical learning stack in which attention, expert routing, and residual mixing share a normalized manifold, yielding sharper subcellular boundaries and improved gradient stability.

\subsection{Problem Setup and Notation}

We address voxelwise subcellular structure segmentation from 3D brightfield microscopy volumes. Given an input $x \in \mathbb{R}^{B\times C\times D\times H\times W}$—batch size $B$, channels $C$, axial depth $D$, in-plane dimensions $(H\times W)$—the goal is to predict a binary mask
$y \in \{0,1\}^{B\times 1\times D\times H\times W}$.
We partition the volume into anisotropic patches of size
$P=(p_z,p_y,p_x)=(2,16,16)$.
After PSF-aware anti-aliasing and striding along $z$, and tiling in $xy$, the token lattice has spatial dimensions
$D' = D/p_z$, $H' = H/p_y$, $W' = W/p_x$,
with $N = D'H'W'$ tokens. Each patch is projected to a hidden dimension $h=512$, yielding
$T \in \mathbb{R}^{B\times N\times h}$.
Divisibility constraints $(D\bmod p_z = H\bmod p_y = W\bmod p_x = 0)$ are satisfied via boundary cropping.

The backbone is a 12-layer normalized Transformer with eight attention heads and a Soft Mixture-of-Experts (MoE) module (72 experts, expansion factor 8) replacing the standard MLP; dropout is $0.1$. All hidden states are $L_2$-normalized, so each block performs rotations on the unit hypersphere rather than unconstrained rescalings. The decoder follows a UNETR-style pyramid~\cite{hatamizadeh2022unetr} with base feature size 64, reshaping intermediate token maps back to $(D'\!\times\!H'\!\times\!W')$ and progressively upsampling to full resolution.

Patch parameters were empirically optimized for brightfield data. Brightfield structure is information-rich in-plane but comparatively sparse along $z$. Setting $p_z=2$ preserves thin axial features, while $p_y=p_x=16$ aggregates local texture into manageable tokens, balancing aliasing suppression and computational load.

\begin{figure}
    \centering
    \includegraphics[width=1\linewidth]{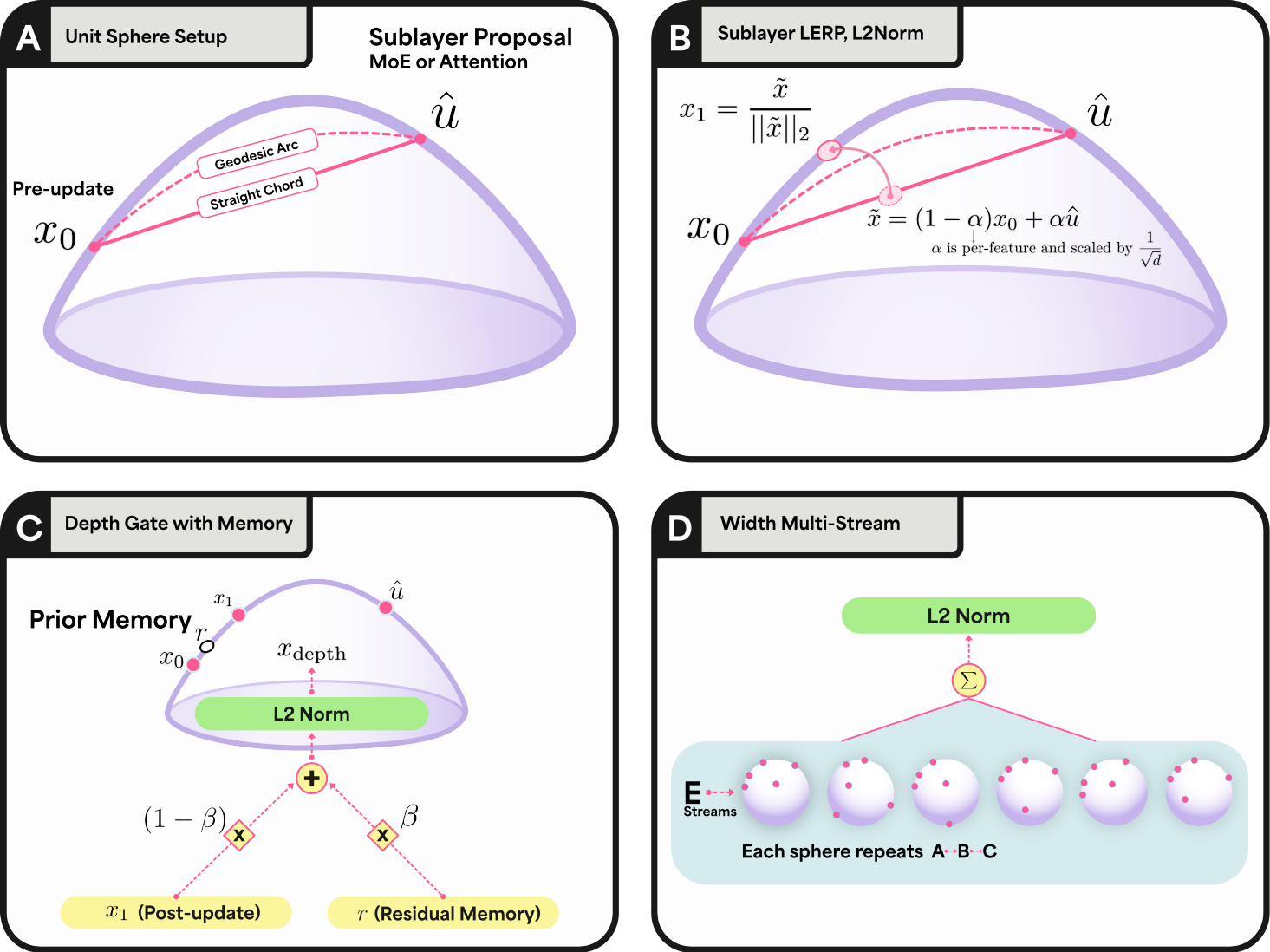}
    \caption{\textit{Hypersphere Setup.} \textbf{[A]} The pre-update $x_0$ and proposal $\hat{u}$ live on the unit sphere. We use LERP + $L_2$ normalization (chord + radial projection), approximating SLERP for small angles. \textbf{[B]} Sublayer outputs from MoE or Attention take small, feature-wise steps that remain stable on the hypersphere. \textbf{[C]} A two-input mixer with gate $\beta$ blends in $\mathbb{R}^d$; the depth gate arbitrates new evidence versus prior memory per feature. \textbf{[D]} $E$ streams are processed in parallel, then reduced and normalized. Intuitively, width = multiple latent opinions, merge = consensus on the hypersphere.}
    \label{fig:hypersphere}
\end{figure}

\subsection{Anisotropic Patch Embedding}

We tokenize 3D brightfield volumes using an anisotropy-aware patch embedding aligned with the PSF. Given $x \in \mathbb{R}^{B\times C\times D\times H\times W}$, we first apply a depthwise 3D convolution with kernel $(p_z,1,1)$, stride $(p_z,1,1)$, and groups $=C$ to low-pass and subsample only along $z$, producing
$x' \in \mathbb{R}^{B\times C\times D'\times H\times W}$,
where $D' = D/p_z$. This PSF-aware anti-aliasing preserves thin axial structures before decimation.

We then fold non-overlapping $xy$ tiles of size $(p_y,p_x)$ into token channels:
$
\text{rearrange}(x'):\! (B,C,D',H,W) \rightarrow (B,D',H',W',p_y p_x C),
$
where $H' = H/p_y$ and $W' = W/p_x$. Each token vector in $\mathbb{R}^{p_y p_x C}$ is mapped to a latent dimension $h$ by a bias-free linear layer, yielding a 3D token lattice in $\mathbb{R}^{B\times D'\times H'\times W'\times h}$.

Crucially, $p_z$ does not appear in the token dimension, since axial downsampling is handled entirely by the anti-alias convolution. This preserves precise spatial indexing and makes the module a drop-in replacement for standard \texttt{rearrange}+linear patchifiers. The three steps—anti-alias + stride in $z$, tile in $xy$, project to $h$—minimize aliasing, respect confocal anisotropy, and produce geometry-faithful tokens for downstream 3D attention and UNETR decoding.

\begin{figure}
    \centering
    \includegraphics[width=0.7\linewidth]{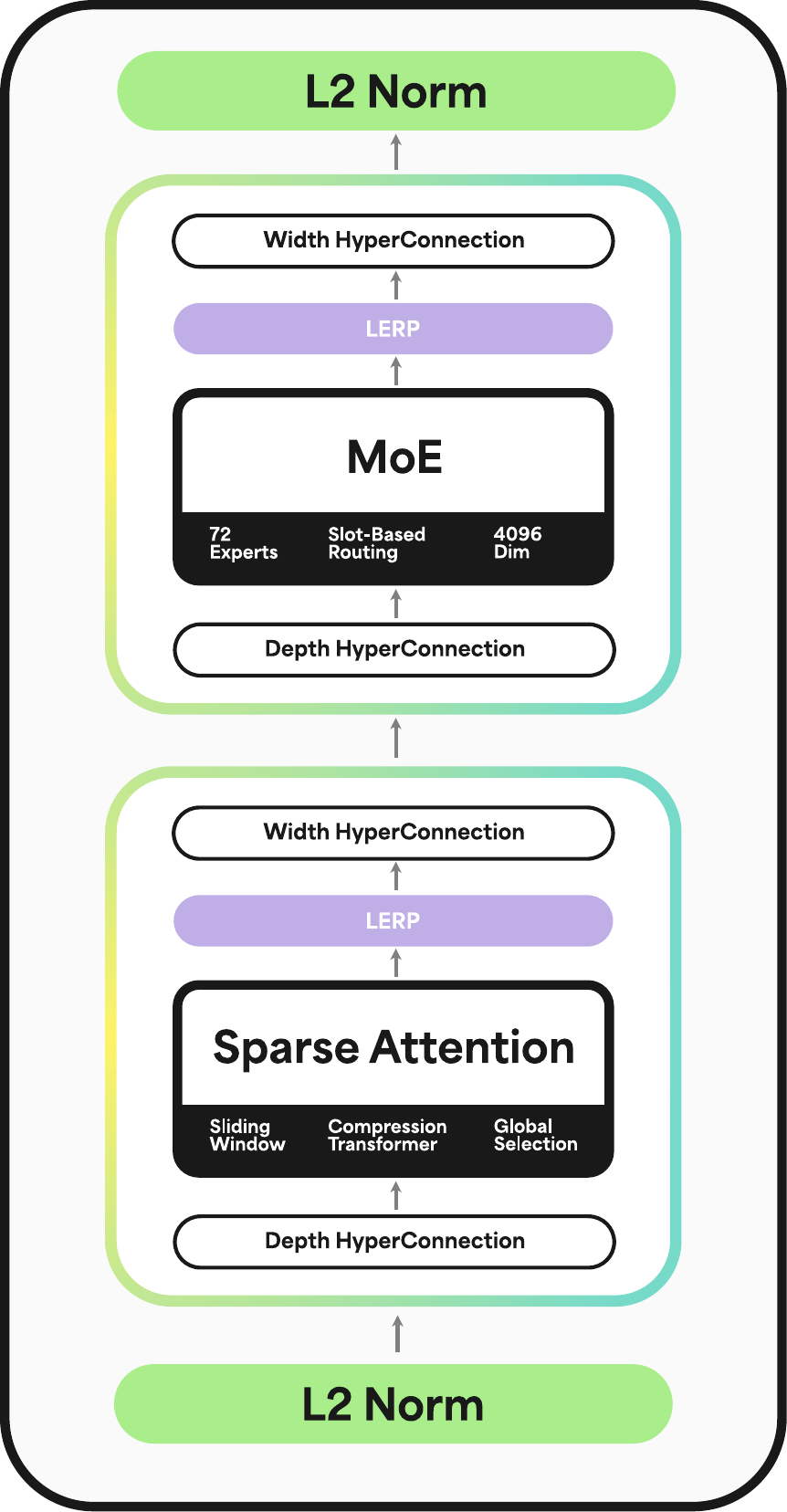}
    \caption{\textbf{Bright 4B Architecture.} Each block follows PreNorm-Attention-Residual and PreNorm-FFN-Residual but with two additions: a cross-stream mixer that blends $E$ residual streams, and a spherical residual gate that applies unit-norm LERP updates. We replace MHA with Native Sparse Attention (sliding window, compressed memory, global selection) and the MLP with SoftMoE (slot-based soft routing + load balancing). Shapes are preserved $(B,N,d)$ throughout.}
    \label{fig:placeholder}
\end{figure}

\subsection{Native Sparse Attention on 3D Token Lattices}

We instantiate Native Sparse Attention (NSA)~\cite{yuan2025native} on vision tokens produced by the anisotropic patch embed. Let $\mathbf{H}\!\in\!\mathbb{R}^{m\times d}$ be the token matrix ($m$ tokens, $d$ channels) and $(\mathbf{Q},\mathbf{K},\mathbf{V})$ the standard projections. Rather than attending to all $\mathbf{K},\mathbf{V}$, NSA \emph{remaps} them into a compact, information-dense set conditioned on the query and a contextual memory. For a query at position $t$:
\[
\widetilde{\mathbf{K}}_t = f_K(\mathbf{q}_t,\mathbf{k}_{1:t},\mathbf{v}_{1:t}),\quad
\widetilde{\mathbf{V}}_t = f_V(\mathbf{q}_t,\mathbf{k}_{1:t},\mathbf{v}_{1:t}),
\]
\[
\mathbf{o}_t^{\star} = \mathrm{Attn}\!\big(\mathbf{q}_t,\widetilde{\mathbf{K}}_t,\widetilde{\mathbf{V}}_t\big).
\]

We realize $f_K,f_V$ as a \emph{mixture of three mapping strategies}
$\mathcal{C}\!=\!\{\text{cmp},\text{slc},\text{win}\}$:
(i) \textbf{compression} (cmp) that aggregates sequential blocks into block-level keys/values,
(ii) \textbf{selection} (slc) that expands a small set of high-salience blocks back to token resolution, and
(iii) a \textbf{sliding window} (win) preserving a local neighborhood. A per-query gate $\mathbf{g}_t\!\in\![0,1]^3$ (sigmoid MLP on $\mathbf{q}_t$) mixes the three routes. This yields a sparse set of remapped keys and sub-quadratic complexity.

\paragraph{Token compression.}
The compression path supplies denoised, long-range context without importing background clutter or inflating compute. Thousands of past tokens are replaced by a compact set of semantic anchors—coarse tiles corresponding to recurring structures such as nuclear or organelle clusters—computed by a lightweight compression transformer $\phi$ (3 layers, 256 dim, $k_v$ heads). All compressed states lie on the same unit hypersphere as local tokens, ensuring scale-stable downstream attention.

We partition past keys/values into blocks of length $B$ with stride $S$ and map each block to one compressed key/value:
\[
\widetilde{\mathbf{K}}^{\text{cmp}}_t = \big\{\phi_K(\mathbf{K}_{i:i+B})\ \big|\ i \in \{1,S{+}1,2S{+}1,\dots\}\big\},
\]
\[
\widetilde{\mathbf{V}}^{\text{cmp}}_t = \big\{\phi_V(\mathbf{V}_{i:i+B})\big\}.
\]
Short-context self-attention with Pre-Norm preferentially preserves patterns consistent across the block (nuclear clusters, organelle fields) while suppressing high-frequency brightfield speckle.

\paragraph{Token selection.}
Given compressed logits $\mathbf{s}_t\!\in\!\mathbb{R}^{\lceil t/S\rceil}$, obtained as softmaxed dot products between $\mathbf{q}_t$ and $\widetilde{\mathbf{K}}^{\text{cmp}}_t$, we select $\kappa$ top blocks:
\[
J_t = \text{TopK}(\mathbf{s}_t,\kappa),
\]
and expand them back to tokens with block size $L$:
\[
\widetilde{\mathbf{K}}^{\text{slc}}_t = \text{concat}\big\{\mathbf{K}_{J:L} \mid J \in J_t\big\},\quad
\widetilde{\mathbf{V}}^{\text{slc}}_t = \text{concat}\big\{\mathbf{V}_{J:L} \mid J \in J_t\big\}.
\]
This path yields high-fidelity long-range evidence only where needed (e.g., elongated nuclei, mitotic events), avoiding quadratic cost.

\paragraph{Sliding window.}
We also preserve a fixed local neighborhood of radius $W$ around $t$:
\[
\widetilde{\mathbf{K}}^{\text{win}}_t = \mathbf{K}_{t-W:t+W},\quad
\widetilde{\mathbf{V}}^{\text{win}}_t = \mathbf{V}_{t-W:t+W},
\]
with $W = 256$ tokens. This route stabilizes gradients, captures boundaries and phase edges, and acts as a strong local prior for brightfield.

\paragraph{Grouped queries and normalization.}
We employ grouped-query attention~\cite{ainslie-etal-2023-gqa} ($k_v$ heads $\le$ heads) so multiple query heads share the same remapped $\widetilde{\mathbf{K}},\widetilde{\mathbf{V}}$, reducing memory while encouraging consistent global decisions across heads. The NSA output $\mathbf{o}_t^{\star}$ is $L_2$-normalized and combined via a diagonal spherical LERP update~\cite{shoemake1985animating}, preventing any branch from winning by scale and improving mixed-precision stability.

\paragraph{Why NSA fits brightfield.}
Brightfield volumes exhibit $z$-anisotropy (PSF blur), halos, and clutter. NSA’s three routes map cleanly to these phenomena: the sliding window preserves crisp edges and near-field interference; the compression pathway aggregates blurry but informative mesoscale context across depth (a semantic low-pass); and the selection pathway restores only distant high-fidelity regions judged salient by the query, sidestepping background noise. Our parameters $(B{=}32,\ S{=}32\ \text{main}/16\ \text{ablation},\ L{=}64,\ \kappa{=}4,\ W{=}128)$ match typical subcellular extents and spacing on our token lattice, yielding a favorable compute–accuracy trade-off. Empirically, NSA reduces block artifacts and recovers sharper, depth-consistent boundaries versus dense attention or window-only variants, while remaining tractable for 3D.

\subsection{Soft Mixture of Experts}

We replace the feed-forward block with a \emph{Soft MoE} layer~\cite{puigcerver2023sparse} that grows capacity without increasing per-token compute. Let $\mathbf{X}\in\mathbb{R}^{m\times d}$ be the token matrix ($m$ tokens, $d$ channels). The layer owns $n$ experts, each applied to $p$ \emph{slots}, for $s=n\cdot p$ slots in total. A learned key matrix $\boldsymbol{\Phi}\in\mathbb{R}^{d\times s}$ produces routing logits; a column-wise softmax yields \emph{dispatch} weights $\mathbf{D}\in\mathbb{R}^{m\times s}$ that convex-combine tokens into slots, $\tilde{\mathbf{X}} = \mathbf{D}^\top \mathbf{X}$. Each expert $f_k:\mathbb{R}^{d}\!\rightarrow\!\mathbb{R}^{d}$ processes its slots, producing $\tilde{\mathbf{Y}}\in\mathbb{R}^{s\times d}$. A row-wise softmax over the same logits produces \emph{combine} weights $\mathbf{C}\in\mathbb{R}^{m\times s}$, and tokens are reconstructed as $\mathbf{Y} = \mathbf{C}\tilde{\mathbf{Y}}$.

Dispatch and combine are both convex: slots are weighted averages of tokens, and tokens are weighted averages of expert outputs. All vectors are unit-normalized, so experts operate on the hypersphere and cannot solve the task by inflating feature scales.

\paragraph{Why MoE helps brightfield.}
Brightfield volumes exhibit high heterogeneity: weak edges, phase halos, uneven illumination, and morphology coupling distant voxels. Soft MoE addresses this in three ways. (i) \emph{Noise-robust aggregation:} the slotting $\tilde{\mathbf{X}}$ performs adaptive, content-aware averaging before expertise is applied, suppressing speckle and halos while preserving coherent structure. (ii) \emph{Specialization without dead experts:} soft routing and gradients through $\mathbf{D},\mathbf{C}$ encourage experts to specialize (thin rims, dense chromatin, defocused planes) without brittle top-$k$ gating. (iii) \emph{Capacity at fixed compute:} setting $s$ to match a dense MLP keeps FLOPs per token unchanged while increasing representational diversity, which is crucial for long 3D sequences where attention already dominates memory. NSA supplies local/coarse/global context; SoftMoE converts these features into task-specific subroutines—denoising, edge sharpening, morphology reasoning—on the unit hypersphere, yielding accurate brightfield-only segmentations.

\subsection{Dynamic Hyper-Connections}
\label{sec:hc}

In a 4B-parameter backbone mixing NSA and SoftMoE, simple residual wiring~\cite{he2016deep} is brittle: one route (e.g., the local window) can dominate early, experts can collapse, and scale drift destabilizes optimization—especially in half precision. \emph{Dynamic Hyper-Connections} (DHC)~\cite{Zhu2024HyperConnections} address this by (i) running $E$ parallel residual streams that allow distinct computations to specialize (local, coarse, global/expert), and (ii) learning input-conditioned \emph{width} and \emph{depth} gates that arbitrate among streams while preserving unit-norm geometry. Intuitively, DHC separates “what to combine” (width mixing across streams) from “how far to move” (depth interpolation on the hypersphere), preventing any single branch from winning by scale and providing depth without gradient brittleness.

\begin{figure}
    \centering
    \includegraphics[width=1\linewidth]{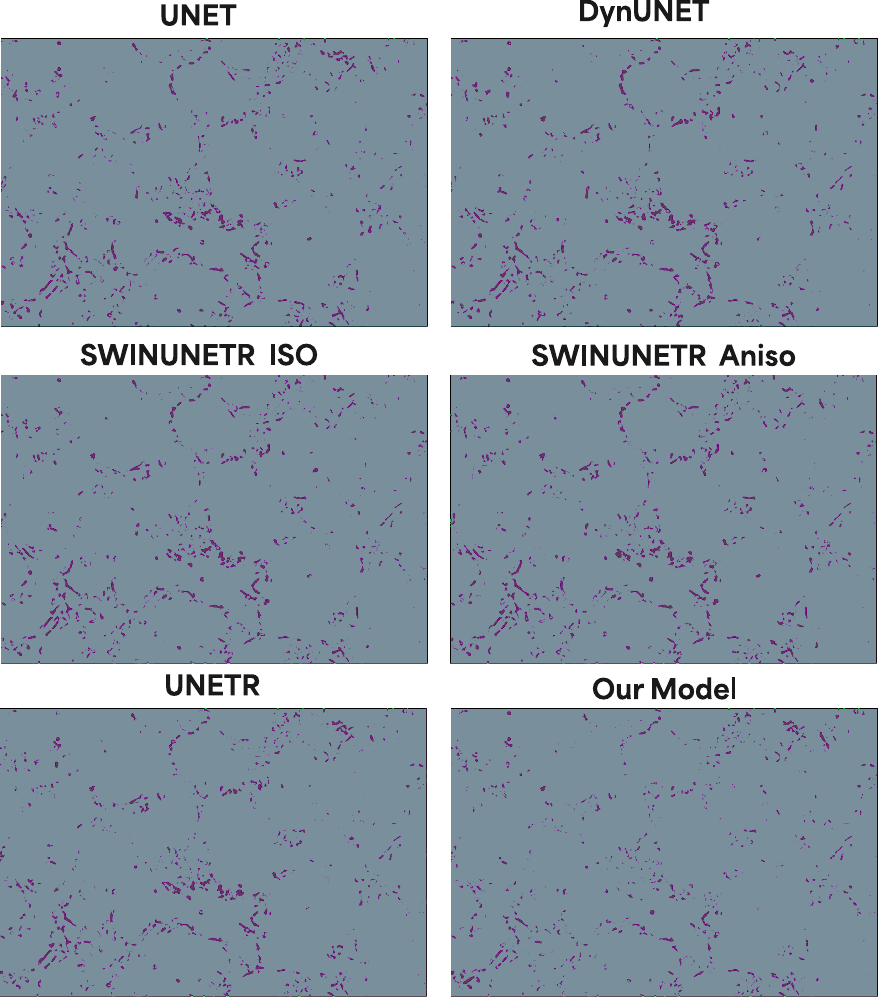}
    \caption{For each model, we visualize the voxel-wise difference between the pseudolabel and the predicted mask.}
    \label{fig:placeholder}
\end{figure}

\begin{figure*}[!t]
    \centering
    \includegraphics[width=0.95\linewidth]{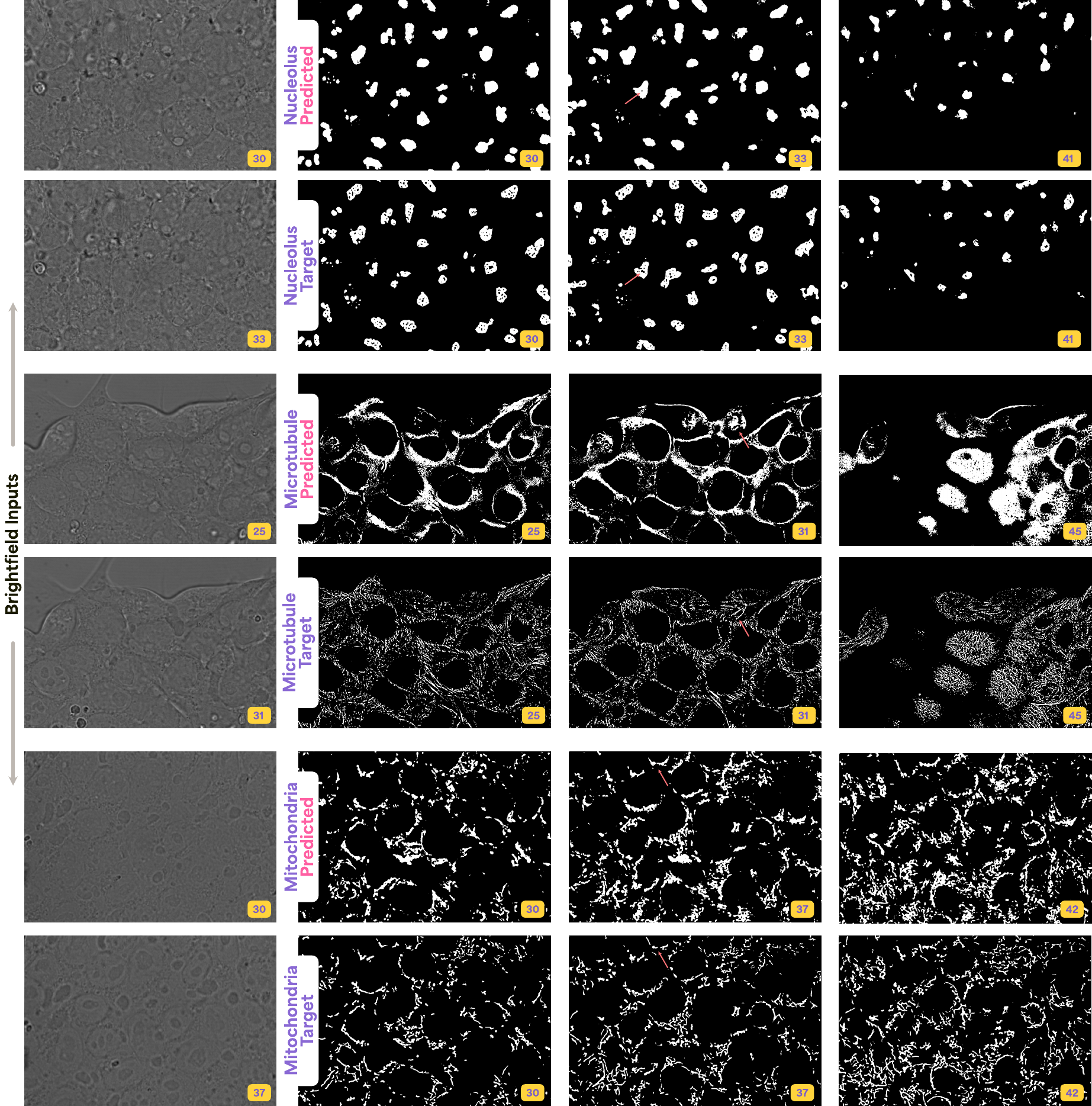}
    \caption{\textbf{Qualitative Segmentation Results for Bright 4B}. The first column shows a representative 2D slice from the 3D bright-field input. The remaining columns display our model's segmentation predictions at three distinct $z$-slices. The $z$-slice index is indicated in the yellow box, with lower indices corresponding to the bottom of the cell and higher indices to the top. Image width is approximately $97\,\mu\text{m}$.}
    \label{fig:placeholder}
\end{figure*}

\section{Experimental Results}

\paragraph{Dataset} We evaluate our method on the publicly available dataset that comprises over 18,000 3D images of live, human induced pluripotent stem cells (hiPSCs, line WTC-11) spanning 25 fluorescently tagged cellular structures ~\cite{viana2023integrated}. Cells were imaged in tightly-packed epithelial-like colonies using spinning-disk confocal microscopy, with each image containing channels for the tagged structure, cell membrane, and DNA. Further, to enable quantitative image and data analyses, accurate and robust segmentations for all 25 intracellular 3D structure localization patterns were created using the Allen Cell \& Structure Segmenter \cite{Chen491035}. In this work, we have used 5 structures for evaluation: 1)~nucleioli-GC tagged via NPM1, 2)~mitochondria tagged via TOMM20, 3)~microtubules tagged via TUBA1B, 4)~DNA labeled with DNA dye, and 5)~Golgi tagged via ST6GAL1. The 3D images used here have size $924\times624\times65$ or $900\times600\times70$ in the $x$, $y$, and $z$ with pixel size of $0.1083\mu m$ and z-spacing of  $0.29\mu m$.

\subsection{Nucleoli}

We first tested our methodology on nucleoli, an organelle chosen because its fluorescent signal is known to be highly predictable from bright-field \cite{ounkomol2018label}. Our results confirm that the model can successfully predict both the location and the overall morphology of individual nucleoli. A key finding is that this predictive accuracy is robust and consistent across the entire z-stack. We did identify a limitation in resolving fine, complex features: the model, for example, tends to miss small internal ``holes" (or vacuoles) that are visible in the ground-truth segmentations.

\subsection{Mitochondria}

We next evaluated our model on mitochondria, which we selected as an ideal intermediate validation target. This organelle poses a distinct challenge: its refractive index is significantly lower than that of nucleoli, resulting in a low-contrast signal, yet its tubular structures remain resolvable (i.e., above the diffraction limit).

Our model demonstrates high fidelity in this low-signal regime. It not only predicts the 3D location of the mitochondrial network—with high consistency at peripheral z-slices (top and bottom of the cell)—but also accurately resolves the orientation of individual tubules in sparse regions. This ability to capture fine-grained orientation from a convolved, low-contrast input strongly suggests the model is implicitly learning to invert the optical path and perform an effective deconvolution of the bright-field point spread function (PSF).

\subsection{Microtubules}

Microtubules, as 25nm sub-diffraction-limit structures, are physically invisible in bright-field microscopy, thus representing the ultimate zero-signal challenge for our model. As hypothesized, the model is unable to reconstruct the fine-grained microtubule network. However, the analysis of this failure case provides a critical insight: rather than producing random noise, the model learns a powerful structural prior.

We observe that its (failed) predictions are correctly confined to the cytoplasmic space and are strictly excluded from the nucleus. This demonstrates that in the total absence of a target signal, the model defaults to learning the valid spatial context of the task. One might hypothesize that microtubule bundles, which are larger and sometimes visible in bright-field (\eg, the mitotic spindle), would be reconstructed. However, our model also fails to resolve these structures in mitotic cells. We attribute this failure to extreme class imbalance: mitotic cells represent only ~$5\%$ of our training data, making them heavily undersampled. The model is thus not exposed to sufficient examples to learn this distinct morphological state.

\begin{table}
\resizebox{\linewidth}{!}{%
\begin{tabular}{@{}lllll@{}}
\toprule
Architecture          & \# Param & IMF &  &  \\ \midrule
UNET                  & 438M     & 20GB      &  &  \\
Dynamic UNET          & \ws{91M}      & 51GB      &  &  \\
UNETR                 & 211M     & 81GB          &  &  \\ 
SWIN UNETR (ISO)      & 248M     & 110GB          &  &  \\
SWIN UNETR (ANISO)    & 247M     & 97GB         &  &  \\
Bright 4B (Our Model) & 4B       & \ws{26GB}      &  & 
\\ \midrule
\end{tabular}%
}
\caption{For each model, we show the empirically chosen parameter counts and inference memory footprint (IMF) that gave the best results for each architecture. Because we use a hardware aligned sparse attention, each query attends only to a bounded set instead of all $N$ tokens, as seen in the UNETR and SWINUNETR variants. Additionally, our anisotropic 3D patch embedding strides early along $z$, hence we never materialize the dense volumes, tokens are compact from the start.}
\label{tab:arch}
\end{table}

\paragraph{Architectural Benchmarks}

We chose the number of parameters for each model based on empirical testing to make the comparison between our four billion parameter model as fair as possible. These parameter counts were the most effective for learning the mitochondria structure, and we found that anything below or above these either degraded the performance or added more compute with little to no gain in segmentation accuracy. More precisely, we noticed that for the base UNET, adding more parameters produced more blobby structures and focus shifted more on overall shape, while too few parameters learned the inherent brightfield noise. The transformer-based models produced reasonable results all within the $\approx 200M$ range. One of our core innovations is seen in Table \ref{tab:arch}, wherein our model despite being four billion parameters only occupies an inference memory footprint of 26GB, which is several scales less than the dense networks. All models were tested with a batch size of one, and all image and patch sizes were the same across all models.

\section{Conclusion}

We presented Bright 4B, a 4-billion parameter foundation model for 3D bright-field segmentation that operates on the unit hypersphere. We designed an anisotropy-aware patch embedding module that aligns with the microscope's PSF. We demonstrated that this approach yields segmentations with high $z$-axis consistency and accurate organelle localization, particularly for structures with strong optical signals (\eg, high refractive index or supra-diffraction-limit size). We provided a solution for scaling model capacity and stable training for the billion parameter scale, while keeping the model's inference memory footprint feasible to run on a single GPU. While this model was only tested for brightfield data, we believe it can be quickly adapted for other scientific domains.

The clear next step is to move beyond pixel-level validation to quantitative biological analysis. Future work must rigorously assess the model's fidelity in downstream tasks, such as measuring nucleolar volume or mitochondrial tubule length.

Furthermore, we found that even in zero-signal regimes like microtubules, the model learns a powerful spatial prior, correctly segmenting the valid cytoplasmic space. Exploring the utility of this implicit nuclear and cellular boundary detection is, in itself, a promising new research direction.

\clearpage
\setcounter{page}{1}

\twocolumn[{%
\renewcommand\twocolumn[1][]{#1}%
\maketitlesupplementary
\begin{center}
    \centering
    \captionsetup{type=table}
\resizebox{\linewidth}{!}{%
\def\arraystretch{1.15}%
\begin{tabular}{lcccccc}
\toprule
\textbf{Metric} & 
\textbf{UNet} & 
\textbf{DynUNet} & 
\textbf{UNETR} & 
\textbf{SwinUNETR} & 
\textbf{Ours (No MoE)} & 
\textbf{Ours} \\
\midrule
\textbf{Params} & 438M & 91M & 211M & 248M & 171M & 4B \\
\textbf{IMF} & 20GB & 51GB & 81GB & 110GB & 24GB & 26GB \\
\midrule 
\textbf{Hausdorff} & \rd{57.94} & \rd{57.94} & 50.98  & 53.41 & 56.55 & \ws{46.08} \\
\textbf{IoU} & 0.396 & 0.319 & 0.368 &   0.362 & \rd{0.287}  & \ws{0.42} \\
\textbf{Precision} & 0.496 & 0.369 & 0.452 & 0.452 & \rd{0.341} & \ws{0.532} \\  
\textbf{HD95} & 29.53 & \rd{29.54} & 13.84  &  18.72 & 24.01 & \ws{9.86} \\  
\textbf{Dice Score} & 0.566 & 0.482 & 0.538 &    0.531 & \rd{0.445} & \ws{0.591} \\
\textbf{Rel. Volume Error} & 37.11 & \rd{95.92} & 51.92  & 47.13 & 95.11 & \ws{25.52} \\  
\textbf{Avg Surface Dist (mm)} & \rd{6.155} & 6.154 & 2.083 & 2.599 & 3.694 & \ws{1.598} \\ 
\textbf{Component Count Error} & \ws{694.61} & 677.06 & 1233.44 & 697.41 & \rd{2898.16} & 1577.92 \\ 
\bottomrule
\end{tabular}
}
    \captionof{table}{\textbf{Bright 4B vs. SOTA Architectures.} Each model was trained using the Mitochondria Benchmark Dataset for the task of 3D segmentation. All models are native 3D and used the 3D brightfield as input and output the generated 3D mask. We detail all the training parameters below for each of the models. We included a version of our model with no mixture of experts to measure the difference of one proposed change of our architecture. We used UNETR as our baseline for testing whether learning on a hypersphere is needed, as its backbone is a vanilla ViT. Despite being a 4 billion parameter model, one of our core design principles of ``more capacity, but used sparsely and selectively" is shown with only a 26GB inference memory footprint (IMF). }
\end{center}%
}]
\appendix

\begin{tcolorbox}[colback=white,colframe=Crimson,title={Sections},fonttitle=\bfseries]
\begin{itemize}
    \item[\ref{quant}] Quantitative Results
    \item[\ref{ablation}] Ablations
    \item[\ref{sec:pca}] Full PCA Analysis
    \item[\ref{moddet}] Model Details
    \item[\ref{sec:trainingrecipe}] Training Recipe
\end{itemize}
\end{tcolorbox}

\section{Quantitative Results} \label{quant}

\paragraph{Dataset} Our benchmark dataset is the Mitochondria dataset, a subset from the publicly available dataset that comprises over 18,000 3D images of live, human induced pluripotent stem cells (hiPSCs, line WTC-11) spanning 25 fluorescently tagged cellular structures ~\cite{viana2023integrated}. From the 25 structures, we selected the mitochondria as an ideal intermediate validation target. This organelle poses a distinct challenge: its refractive index is significantly lower than that of nucleoli, resulting in a low-contrast signal, yet its tubular structures remain resolvable (i.e., above the diffraction limit). In addition, mitochondrial morphology exhibits a broad spectrum of scales and topologies within the same volume (puncta, short rods, and extended reticular networks), making it a natural test of whether the model has learned genuinely shape-aware, topology-sensitive representations rather than simple blob detectors. Mitochondria also reside in cytoplasmic regions with heterogeneous, textured brightfield backgrounds, providing a stringent probe for each architecture’s ability to leverage long-range context and denoise halo-like phase artifacts without erasing fine structure.

\paragraph{Geometry on the hypersphere beats brute-force convolution.} Pure convolutional neural networks see only local neighborhoods. They can give reasonable per-voxel Dice (0.566 for a base UNET), but they struggle to keep extended morphology consistent, as seen in the high HD95 and average surface distance. UNETR \cite{hatamizadeh2022unetr}, on the other hand, brings dense ViT \cite{dosovitskiy2020image} attention over patches, so it sees the whole volume. This improves surface metrics, going from average surface distance of around six down to two and HD95 from around 30 to 13.8. We attribute this to local convolutions versus global token interactions. We used this finding and found a middle-ground that split the attention into local, compressed, and selected paths \cite{yuan2025native}, so that \emph{long-range context is structured} rather than uniformly dense. And because each layer's tokens live on the unit sphere, updates are small, gated spherical lerps rather than unbounded residual jumps \cite{loshchilov2024ngpt, tongzhouw2020hypersphere, shoemake1985animating,Zhu2024HyperConnections}.

\begin{figure}[t]
    \centering
    \includegraphics[width=1\linewidth]{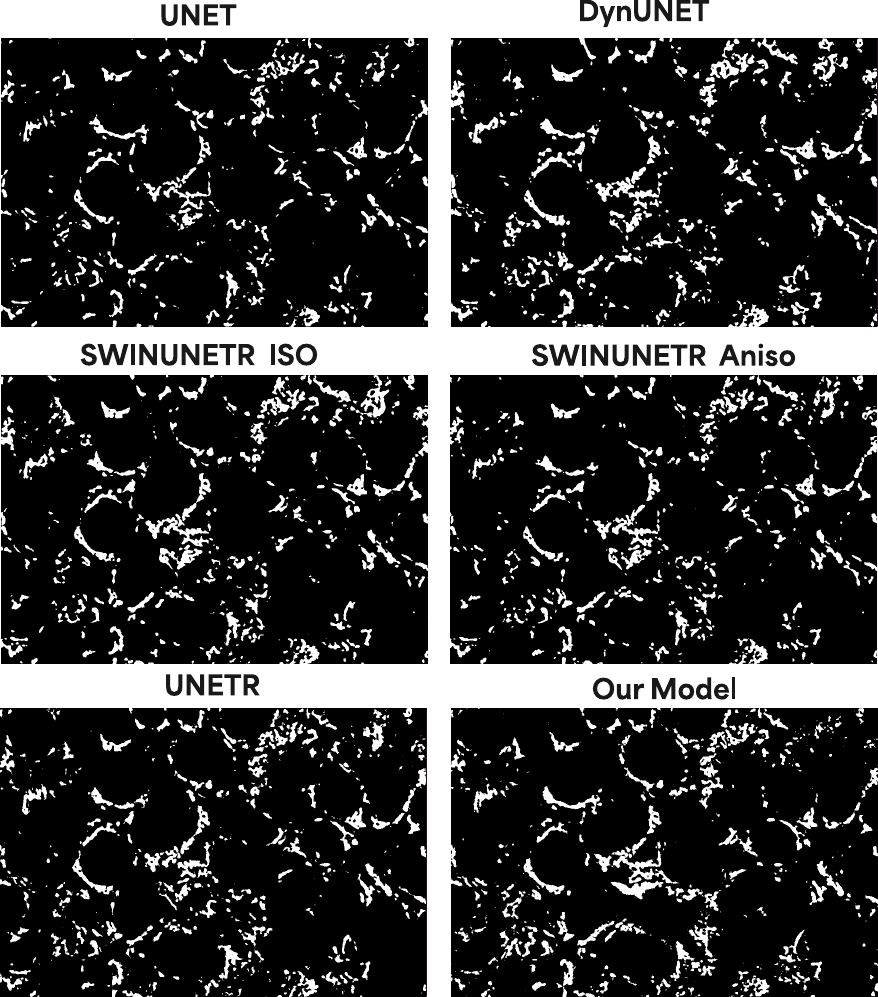}
    \caption{\textit{The Otsu thresholded } outputs for a single slice ($z=32$) for each of the networks we benchmark against. The mod}
    \label{fig:segs-network}
\end{figure}

\begin{figure*}
    \centering
    \includegraphics[width=1\linewidth]{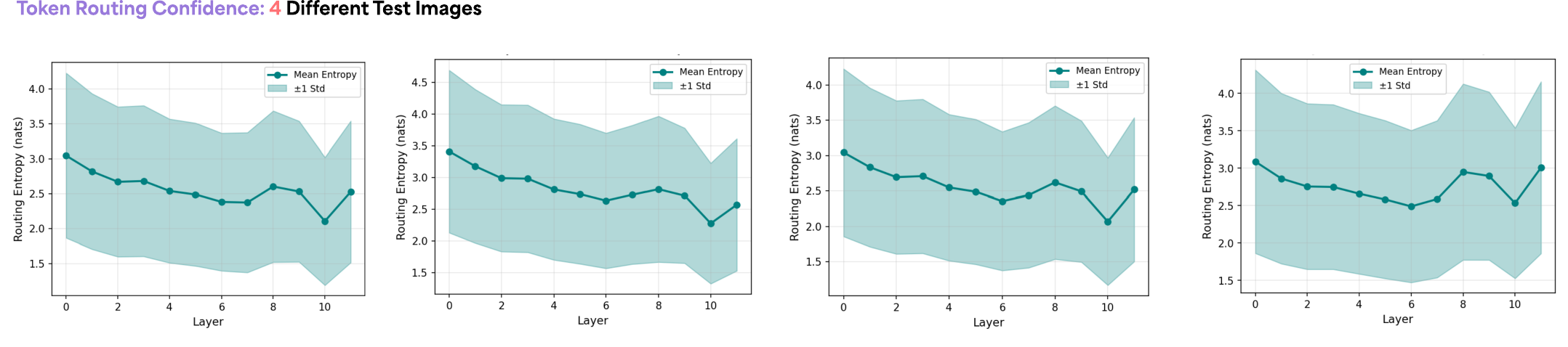}
    \includegraphics[width=0.9\linewidth]{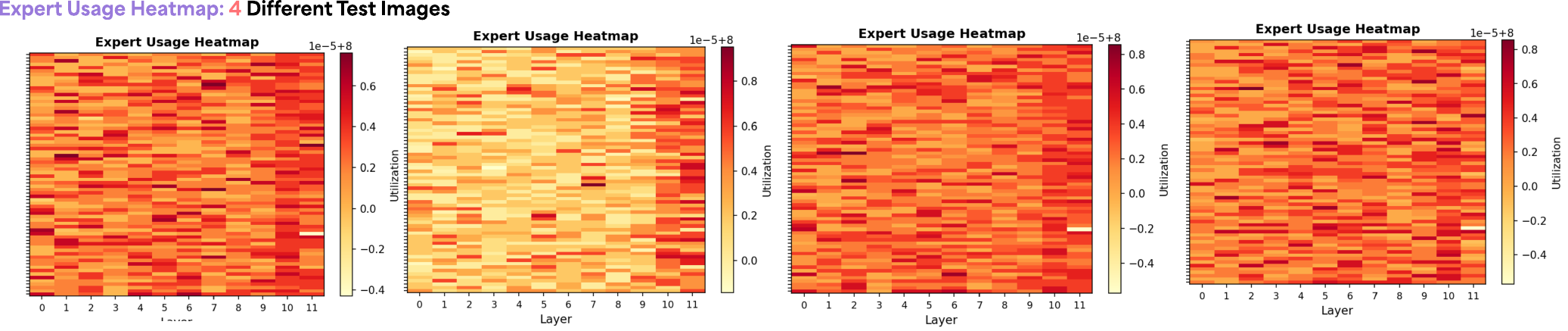}
    \caption{\textbf{Dynamic Slots SoftMoE routing analysis.} \textit{Token Routing Confidence.} Token routing entropy $H = -\sum p \log p$ measuring routing confidence. Here we show how four different test images during inference time route tokens throughout the 12 layers of the network. The general trend seems to be that routing in layer 10 is least confident for our MoE layer. \textit{Expert Usage Heatmap.}  Usage heatmap showing expert specialization across depth for each of the input test images. For image 2, there seems to be very little expert usage until later stages. This is because for Image 2, slices 1-20 are background (pure black) and the last 15 are nearly the same. We show this in Figure \ref{fig:img2}.}
    \label{fig:placeholder}
\end{figure*}

\paragraph{Why MoE Matters} We elaborate on this more in the next section \ref{ablation}, but we emphasize here that it is not enough to have native sparse attention (NSA) and hypersphere geometry. The expert specialization turns the geometry into useful segmentations. The multiple residual streams and the soft mixture of experts (soft-MoE) \cite{puigcerver2023sparse} allows the model to maintain distinct subspaces and let the experts selectively operate in those subspaces \cite{jordan1994hierarchical}. 

\section{Ablations} \label{ablation}

\paragraph{Reducing to One Expert} When we ablate the Soft-MoE from 72 experts down to a single expert, we are effectively collapsing the entire feed-forward part of the backbone back into a single, shared non-specialized MLP, and this breaks several of the design assumptions we rely on for brightfield. In the full model, slot-based soft routing sends different token ``types" (clean nuclear interior, halo edges, cytoplasmic texture, background junk) to different experts, so each expert can carve out a simpler subproblem in feature space (\eg, one expert primarily denoises halos, another sharpens boundaries, another stabilizes faint structures). NSA is then free to mix this very heterogeneous local + global evidence, because the MoE layer downstream has enough conditional capacity to disentangle those mixed signals. With only one expert, all of these heterogeneous contexts must be modeled by a single shared transformation, so gradients from ``noise suppression" tokens and ``boundary sharpening" tokens constantly interfere. This both lowers the effective capacity of the model and makes optimization more brittle, especially on the hypersphere where we rely on small, well-conditioned updates. 

During our experiments, we observed two major failure points---slower convergence and significantly worse masks, particularly in the hard brightfield regimes where expert specialization was originally intended to help. The loss during training would fluctuate between $\pm 0.05$ for at least two to three epochs before making a noticeable decision, hence making the loss curve jagged compared to the smooth decisive decline in our 72 expert model. Additionally, we found that the loss plateaued fairly quickly after learning the global shape and failed to learn the intricacies of the mitochondria.  

\paragraph{Testing with Fluorescent Tagged Image} We gave our model the fluorescent tagged image of the mitochondria and found the model seemed to reuse some of the same features, or at least show a semblance of similar structure learned in the brightfield. We include the resulting analysis plots for this single Image 1 as well.

\begin{figure}[h]
    \centering
    \includegraphics[width=1\linewidth]{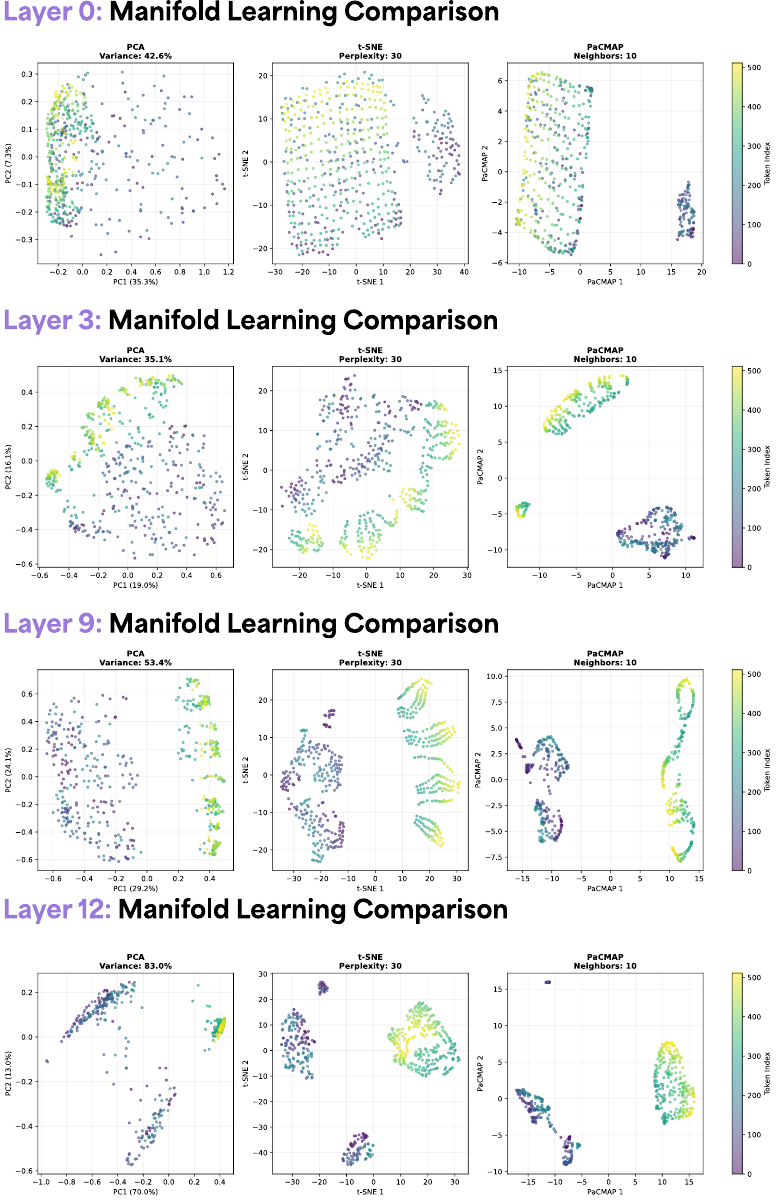}
    \caption{\textit{Image 1.} \textbf{Fluorescent Tagged} We tested what the model would output based on the fluorescent tagged image as input for the mitochondria. We only tested this with Image 1, but found that the model picks up on some of the same learned geometry. Further research is needed to validate how feasible the model can be used as a backbone for contrastive and continual learning. Nonetheless, we make no claims here on the capability and extension to those tasks/domains.  \textbf{ Comparison of dimensionality reduction techniques for layer $\ell$.} (a) PCA: linear projection preserving global variance. (b) t-SNE: nonlinear projection preserving local neighborhoods (perplexity $=30$). (c) PaCMAP: balanced preservation of local and global structure. }
    \label{fig:placeholder}
\end{figure}

\begin{figure*}
    \centering
    \includegraphics[width=0.75\linewidth]{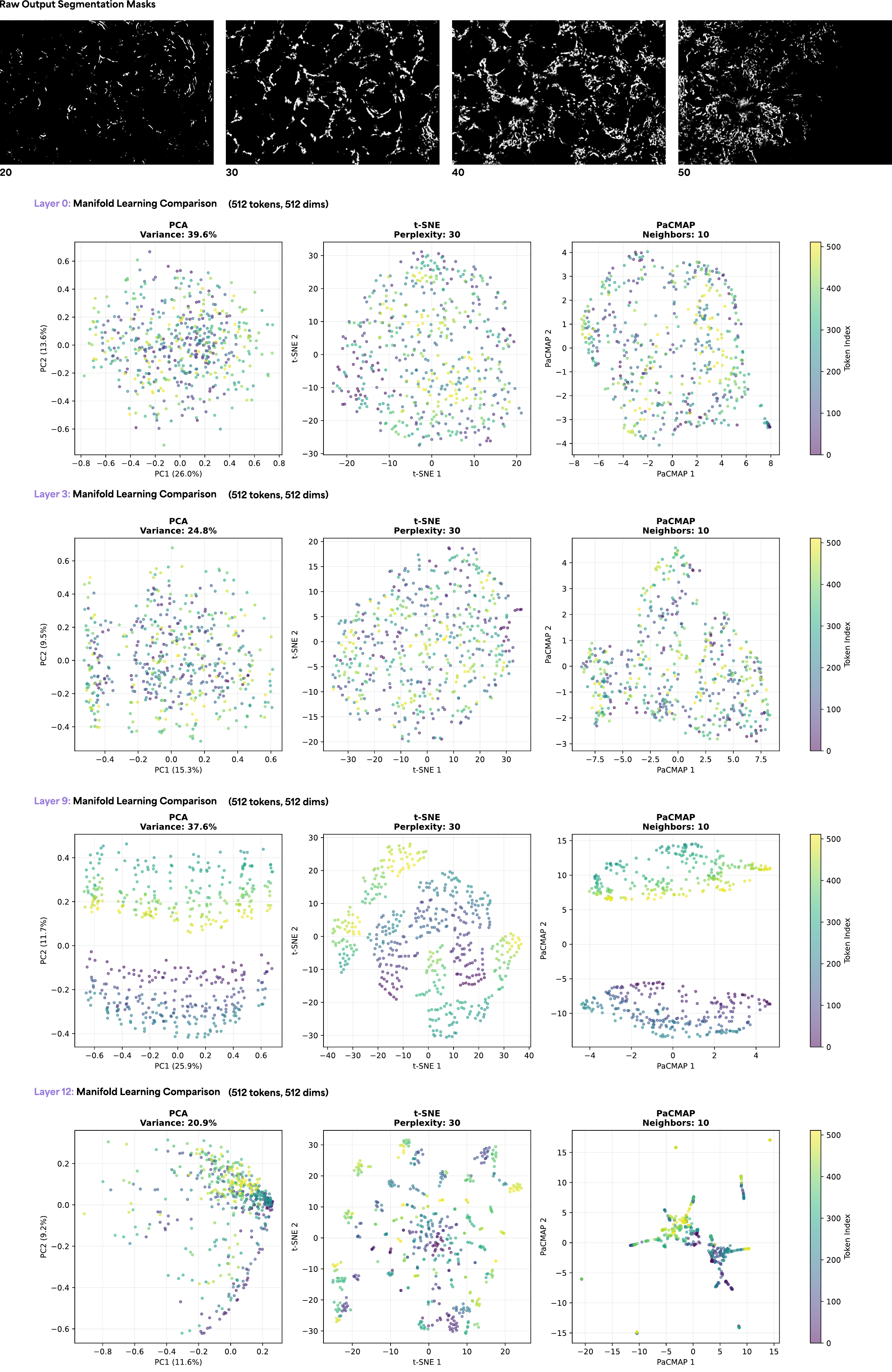}
    \caption{\textit{Image 1.} \textbf{At the top}, we show different slices of the raw predicted output mask, so no thresholding is done to these images. We chose to use the output mask at different slices because it gave a better insight into what the model at different layers were learning. \textbf{ Comparison of dimensionality reduction techniques for layer $\ell$.} (a) PCA: linear projection preserving global variance. (b) t-SNE: nonlinear projection preserving local neighborhoods (perplexity $=30$). (c) PaCMAP: balanced preservation of local and global structure. }
    \label{fig:placeholder}
\end{figure*}

\begin{figure*}
    \centering
    \includegraphics[width=0.75\linewidth]{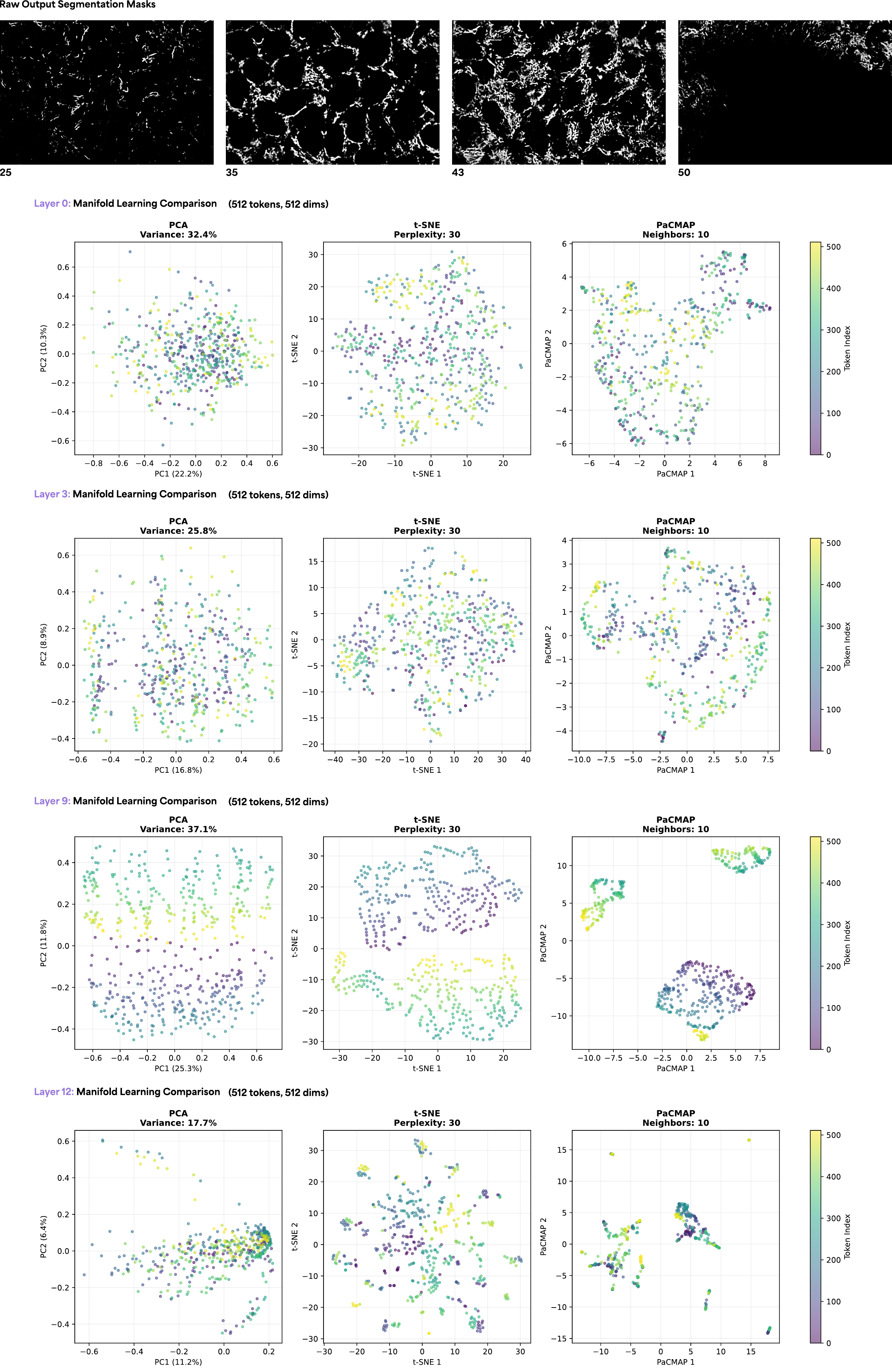}
    \caption{\textit{Image 2.} \textbf{At the top}, we show different slices of the raw predicted output mask, so no thresholding is done to these images. We chose to use the output mask at different slices because it gave a better insight into what the model at different layers were learning.\textbf{ Comparison of dimensionality reduction techniques for layer $\ell$.} (a) PCA: linear projection preserving global variance. (b) t-SNE: nonlinear projection preserving local neighborhoods (perplexity $=30$). (c) PaCMAP: balanced preservation of local and global structure.  }
    \label{fig:img2}
\end{figure*}

\section{Full PCA Analysis} \label{sec:pca}

We provide a full analysis that showcases the representations learned throughout the layers of our model. We use four test images and perform the analysis on each of them to show how our network reacts during inference time to different inputs.

\begin{figure*}[t]
    \centering
        \includegraphics[width=1\linewidth]{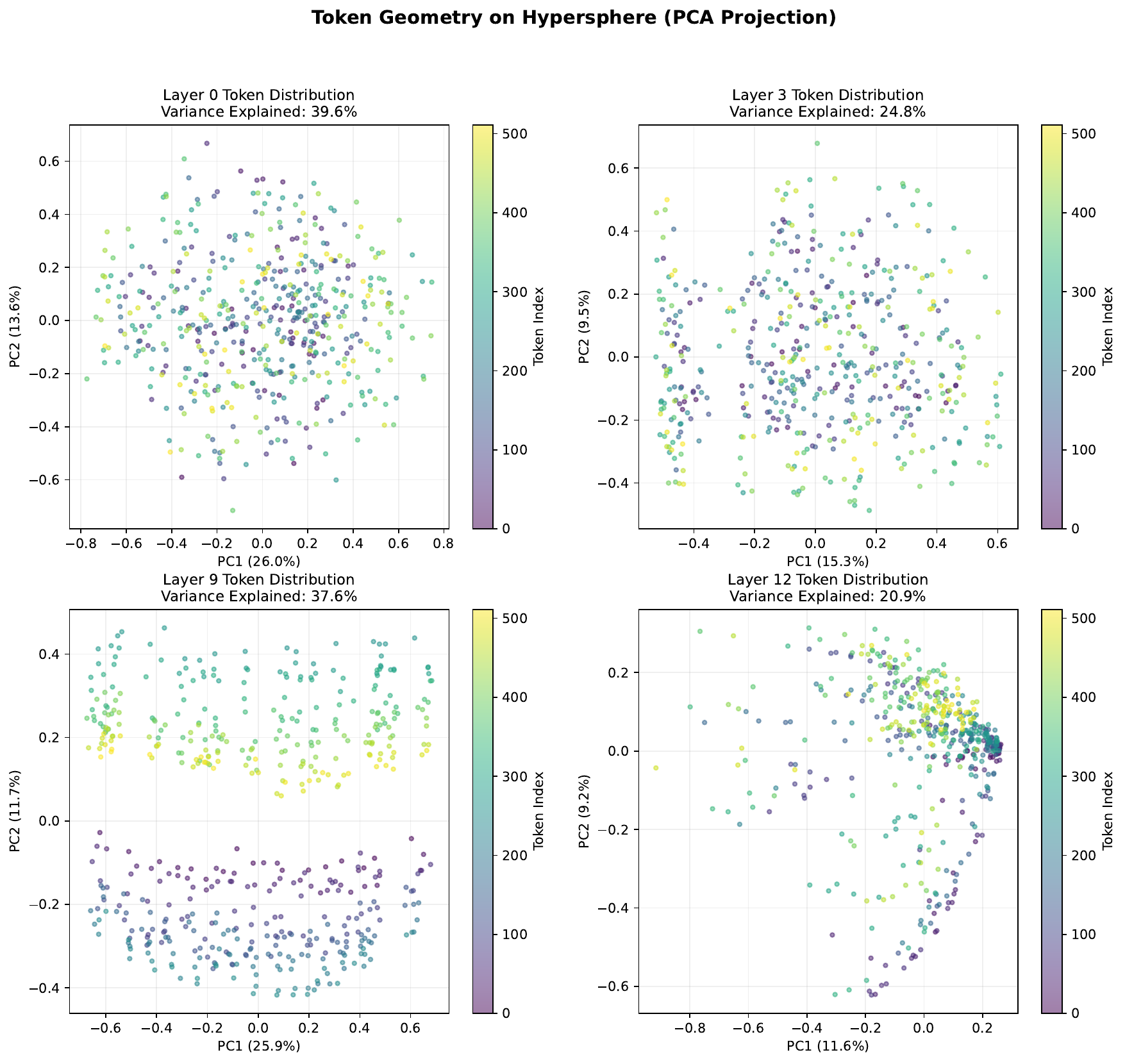}
    \caption{\textit{Image 1 PCA Analysis}}
    \label{fig:placeholder}
\end{figure*}

\begin{figure*}[t]
    \centering
        \includegraphics[width=1\linewidth]{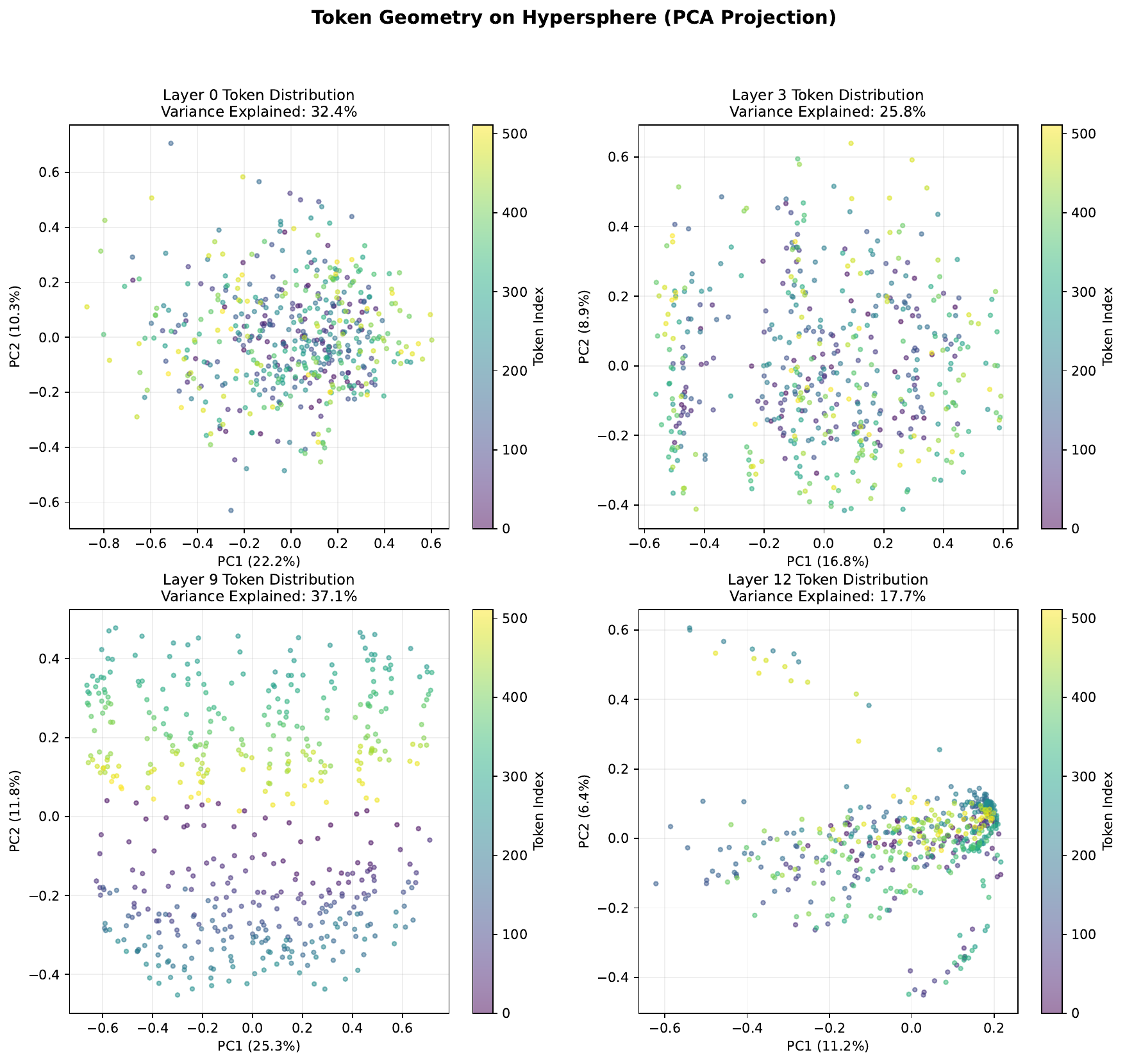}
    \caption{\textit{Image 2 PCA Analysis}}
    \label{fig:placeholder}
\end{figure*}

\begin{figure*}[t]
    \centering
        \includegraphics[width=1\linewidth]{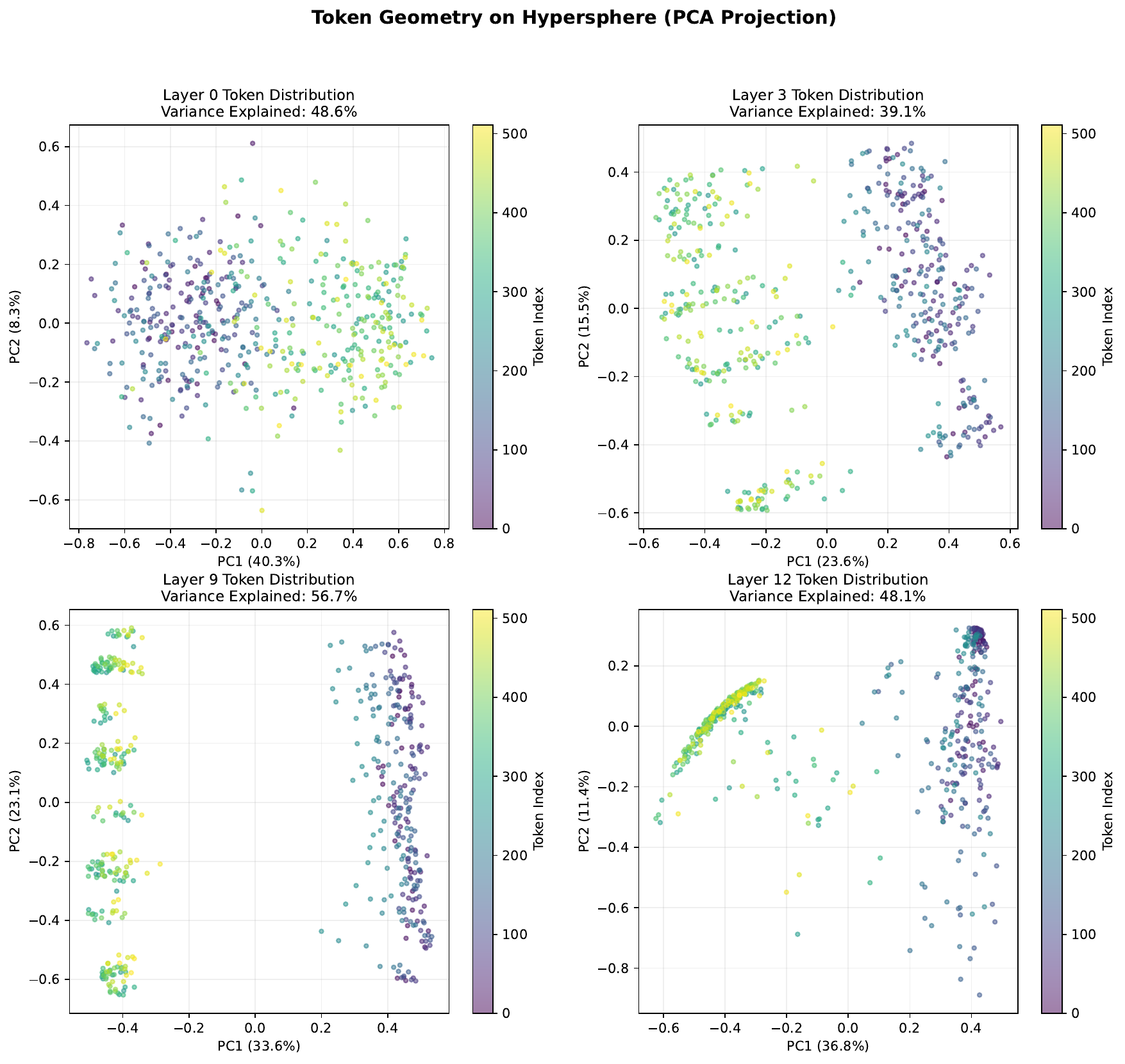}
    \caption{\textit{Image 3 PCA Analysis}}
    \label{fig:placeholder}
\end{figure*}

\begin{figure*}[t]
    \centering
        \includegraphics[width=1\linewidth]{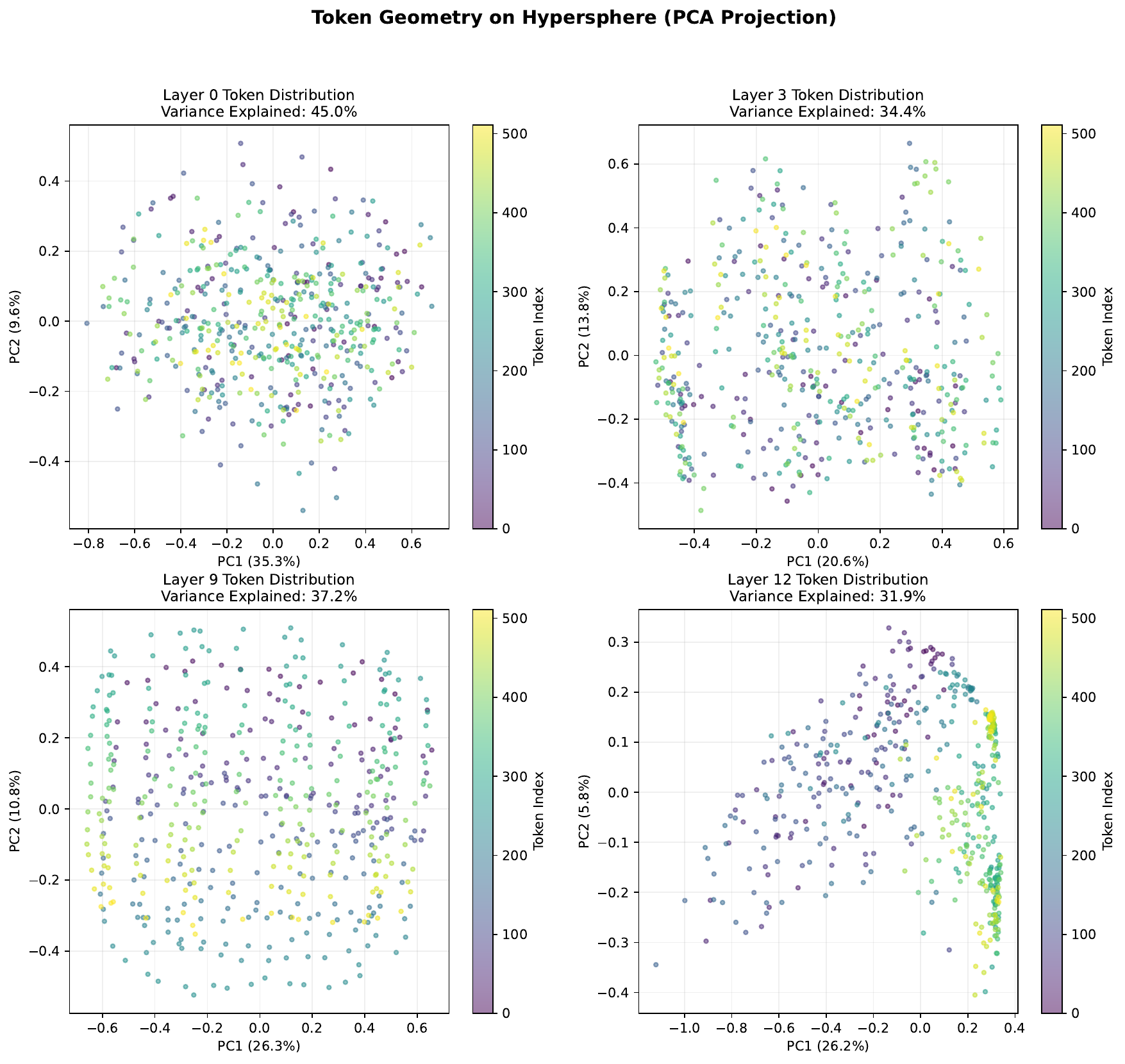}
    \caption{\textit{Image 4 PCA Analysis}}
    \label{fig:placeholder}
\end{figure*}

\section{Model Details} \label{moddet}

Given an input $x \in \mathbb{R}^{B\times C\times D\times H\times W}$—batch size $B$, channels $C$, axial depth $D$, in-plane dimensions $(H\times W)$—the goal is to predict a binary mask
$y \in \{0,1\}^{B\times 1\times D\times H\times W}$.
We partition the volume into anisotropic patches of size
$P=(p_z,p_y,p_x)=(2,16,16)$.
After PSF-aware anti-aliasing and striding along $z$, and tiling in $xy$, the token lattice has spatial dimensions
$D' = D/p_z$, $H' = H/p_y$, $W' = W/p_x$,
with $N = D'H'W'$ tokens. Each patch is projected to a hidden dimension $h=512$, yielding
$T \in \mathbb{R}^{B\times N\times h}$.
Divisibility constraints $(D\bmod p_z = H\bmod p_y = W\bmod p_x = 0)$ are satisfied via boundary cropping.

The backbone is a 12-layer normalized Transformer with eight attention heads and a Normalized Soft Mixture-of-Experts (MoE) module (72 experts, expansion factor 8) replacing the standard MLP; dropout is $0.1$. All hidden states are $L_2$-normalized, so each block performs rotations on the unit hypersphere rather than unconstrained rescalings. The decoder is UNETR~\cite{hatamizadeh2022unetr} with base feature size 64, reshaping intermediate token maps back to $(D'\!\times\!H'\!\times\!W')$ and progressively upsampling to full resolution.

\paragraph{Normalized Depth/Width Residual Formulation.}
Let $\mathbf{H}\!\in\!\mathbb{R}^{m\times d}$ be tokens ($m$ positions, $d$ channels). We maintain $E$ streams, $\mathbf{H}^{(e)}$, and work with unit-norm states $\bar{\mathbf{H}}^{(e)}=\text{Norm}(\mathbf{H}^{(e)})$ as in \cite{loshchilov2024ngpt}. DHC produces (a) a \emph{width mixer} $\mathbf{M}_w(\cdot)\!\in\!\mathbb{R}^{E\times E}$ that linearly combines streams, and (b) a \emph{depth gate} $\boldsymbol{\alpha}(\cdot)\!\in\!\mathbb{R}^{d}_{\ge 0}$ that controls the step size per channel. We use light, input-conditioned parametrizations initialized near identity:
\begin{align}
\bar{\mathbf{H}} &= \big[\,\bar{\mathbf{H}}^{(1)}\,\cdots\,\bar{\mathbf{H}}^{(E)}\,\big]\in\mathbb{R}^{m\times E\times d}, \\
\mathbf{M}_w(\bar{\mathbf{H}}) &= \mathbf{A}_w + s_\alpha\,\tanh\!\big(\textstyle\frac{1}{m}\sum\nolimits_{t}\bar{\mathbf{H}}_{t:}\mathbf{W}_w\big),  \mathbf{A}_w\approx\mathbf{I}_E, \label{eq:hc-width} \\
\boldsymbol{\alpha}(\bar{\mathbf{H}}) &= s_\beta\,\sigma\!\big(\textstyle\frac{1}{m}\sum\nolimits_{t}\bar{\mathbf{H}}_{t:}\mathbf{W}_\beta\big)\in\mathbb{R}^{d}_{\ge 0}, \quad s_\alpha,s_\beta\ll 1. \label{eq:hc-depth}
\end{align}
Width mixing applies $\mathbf{M}_w$ along the stream axis,
\begin{align}
\bar{\mathbf{H}}_{\text{mix}}^{(e)} \;=\; \sum_{e'=1}^{E}\mathbf{M}_w[e,e']\,\bar{\mathbf{H}}^{(e')}, \qquad e=1,\dots,E,\label{eq:hc-mix}
\end{align}
and the core block (NSA $+$ SoftMoE) proposes a normalized update $\widehat{\mathbf{U}}^{(e)}\!=\!\text{Norm}\big(\texttt{NSA}(\bar{\mathbf{H}}_{\text{mix}}^{(e)})+\texttt{SoftMoE}(\bar{\mathbf{H}}_{\text{mix}}^{(e)})\big)$. Each stream is then updated by a spherical LERP (the nGPT approximation to SLERP), gated \emph{per channel}:
\begin{align}
\mathbf{H}^{(e)}_{\text{out}} = \bar{\mathbf{H}}^{(e)} \;+\; \boldsymbol{\alpha}(\bar{\mathbf{H}})\odot\!\big(\widehat{\mathbf{U}}^{(e)}-\bar{\mathbf{H}}^{(e)}\big).\label{eq:hc-slerp}
\end{align}
We finally reduce streams (sum/mean) to produce the block output. With $E{=}1$ and $s_\alpha\!=\!s_\beta\!=\!0$, DHC exactly recovers a Pre-Norm residual, so training starts as a stable baseline and gradually learns to exploit width/depth flexibility. 

\subsection{Token compression}

\begin{align}
\widetilde{\mathbf{K}}^{\text{cmp}}_t &= \Big\{\,\phi_K\!\big(\mathbf{K}_{i:i+B}\big)\Big|\;i\in\{1,S+1,2S+1,\ldots\}\Big\},\\
\widetilde{\mathbf{V}}^{\text{cmp}}_t &= \Big\{\,\phi_V\!\big(\mathbf{V}_{i:i+B}\big)\,\Big\}. \label{eq:cmp-map}
\end{align}

We use $B\!=\!32$, $S\!=\!32$ (no overlap, faster), and in the ablations (50\% overlap) which slightly helps punctate structures.

\paragraph{Sliding window.}
Finally, we preserve a fixed local neighborhood of radius $W$ around $t$,
\begin{align}
\widetilde{\mathbf{K}}^{\text{win}}_t=\mathbf{K}_{t-W:t+W}, \\
\widetilde{\mathbf{V}}^{\text{win}}_t=\mathbf{V}_{t-W:t+W},
\end{align}
with $W\!=\!256$ tokens. This route stabilizes gradients, captures boundaries and phase edges, and acts as a strong prior for brightfield where most cues are local.

\paragraph{Token selection.}
Given compressed logits $\mathbf{s}_t\!\in\!\mathbb{R}^{\lceil t/S\rceil}$ computed as softmaxed dot products between $\mathbf{q}_t$ and $\widetilde{\mathbf{K}}^{\text{cmp}}_t$, we select $\kappa$ top blocks and \emph{expand} them back to tokens with block size $L$ (we use $L\!=\!32$, $\kappa\!=\!8$):
\begin{align}
\mathcal{J}_t = \text{TopK}(\mathbf{s}_t,\kappa),\\
\widetilde{\mathbf{K}}^{\text{slc}}_t = \text{concat}\big\{\mathbf{K}_{J:L}\;|\;J\!\in\!\mathcal{J}_t\big\}, \\
\widetilde{\mathbf{V}}^{\text{slc}}_t = \text{concat}\big\{\mathbf{V}_{J:L}\;|\;J\!\in\!\mathcal{J}_t\big\}. \label{eq:slc-map}
\end{align}
This path yields high-fidelity long-range evidence only where needed (\eg, elongated nuclei, mitotic events), avoiding quadratic cost.

\paragraph{Sliding window.}
Finally, we preserve a fixed local neighborhood of radius $W$ around $t$,
\begin{align}
\widetilde{\mathbf{K}}^{\text{win}}_t=\mathbf{K}_{t-W:t+W}, \\
\widetilde{\mathbf{V}}^{\text{win}}_t=\mathbf{V}_{t-W:t+W},
\end{align}
with $W\!=\!256$ tokens. This route stabilizes gradients, captures boundaries and phase edges, and acts as a strong prior for brightfield where most cues are local.

\subsection{Soft Mixture of Experts}

We replace the feed-forward block with a \emph{Soft MoE} layer \cite{puigcerver2023sparse} that grows capacity without increasing per-token compute. Let the token matrix be $\mathbf{X}\in\mathbb{R}^{m\times d}$ ($m$ tokens, $d$ channels). The layer owns $n$ experts, each applied to $p$ \emph{slots}, for a total of $s=n\!\cdot\!p$ slots. A learned key matrix $\boldsymbol{\Phi}\in\mathbb{R}^{d\times s}$ produces routing logits; a column-wise softmax yields \emph{dispatch} weights $\mathbf{D}\in\mathbb{R}^{m\times s}$ that convex-combine tokens into slots:
\begin{align}
\mathbf{D}_{ij} &= \frac{\exp\!\big((\mathbf{X}\boldsymbol{\Phi})_{ij}\big)}{\sum_{i'=1}^{m}\exp\!\big((\mathbf{X}\boldsymbol{\Phi})_{i'j}\big)}, \quad
\tilde{\mathbf{X}} = \mathbf{D}^\top \mathbf{X}\in\mathbb{R}^{s\times d}. \label{eq:dispatch}
\end{align}
Each expert $f_k:\mathbb{R}^{d}\!\rightarrow\!\mathbb{R}^{d}$ processes its $p$ assigned rows of $\tilde{\mathbf{X}}$, producing slot outputs $\tilde{\mathbf{Y}}\in\mathbb{R}^{s\times d}$. Tokens are then reconstructed by a \emph{combine} step that uses a row-wise softmax over the same logits to form weights $\mathbf{C}\in\mathbb{R}^{m\times s}$:
\begin{align}
\mathbf{C}_{ij} &= \frac{\exp\!\big((\mathbf{X}\boldsymbol{\Phi})_{ij}\big)}{\sum_{j'=1}^{s}\exp\!\big((\mathbf{X}\boldsymbol{\Phi})_{ij'}\big)}, \quad
\mathbf{Y} = \mathbf{C}\,\tilde{\mathbf{Y}}\in\mathbb{R}^{m\times d}. \label{eq:combine}
\end{align}
Both (\ref{eq:dispatch})–(\ref{eq:combine}) are convex: slots are weighted averages of tokens, and tokens are weighted averages of expert outputs. In practice we set $s$ to match the FLOPs of a dense MLP layer; a light load-balancing term encourages high-entropy routing and prevents expert collapse,
\begin{align}
\mathcal{L}_{\text{lb}} = \lambda \big( \underbrace{\text{H}_{\text{row}}(\mathbf{C})}_{\text{balanced combine}} +  \underbrace{\text{H}_{\text{col}}(\mathbf{D})}_{\text{balanced dispatch}} \big),
\end{align}
where $\text{H}$ denotes mean entropy over rows/columns. All vectors are unit-normalized, so experts operate on the hypersphere and cannot solve the task by inflating feature scales.

For a typical field of view $(32\times128\times128)$, this configuration yields $(N=(32/2)\cdot(128/16)\cdot(128/16)=1024)$ tokens. With a hidden size of $h{=}512$, the resulting token tensor $(B\times1024\times512)$ fits comfortably within a single GPU and keeps self-attention complexity sub-quadratic under the applied sparsity.

\section{Training Recipe}
\label{sec:trainingrecipe}

Our backbone uses 512-dimensional patch embeddings with 16 attention heads across 12 Transformer layers. Each block combines Native Sparse Attention with a 72-expert Dynamic Slots SoftMoE layer (expert\_mult$=8$), yielding over 4B parameters while activating only a small subset per token. Two residual streams, implemented via Dynamic Hyper-Connections, provide parallel gradient pathways and width/depth gating, which is critical for stabilizing a 12-layer, sparsely-activated backbone on 3D data. Anisotropic 3D patches of size $[2,16,16]$ are chosen to match confocal microscopy anisotropy: the shallow $z$ extent respects the PSF and axial blur, while larger $xy$ tiles preserve sufficient in-plane context for mitochondrial morphology. A decoder feature size of 64 provides adequate capacity for dense 3D segmentation without dominating the parameter budget, and a dropout rate ($0.1$) helps prevent expert co-adaptation and collapse given the large global capacity. We believe that having all the learning done on the unit sphere was what enabled the training stability in our model. We tested several aspects of our model with $L_2$ normalization omitted at different layers and found instability at the later training stages. The model would converge well, but fail to train further if we did not normalize the depth and width residual connections specifically.

{
    \small
    \bibliographystyle{ieeenat_fullname}
    \bibliography{main}

@article{viana2023integrated,
  title={Integrated intracellular organization and its variations in human iPS cells},
  author={Viana, Matheus P and Chen, Jianxu and Knijnenburg, Theo A and Vasan, Ritvik and Yan, Calysta and Arakaki, Joy E and Bailey, Matte and Berry, Ben and Borensztejn, Antoine and Brown, Eva M and others},
  journal={Nature},
  volume={613},
  number={7943},
  pages={345--354},
  year={2023},
  publisher={Nature Publishing Group UK London}
}

@article{loshchilov2024ngpt,
  title={ngpt: Normalized transformer with representation learning on the hypersphere},
  author={Loshchilov, Ilya and Hsieh, Cheng-Ping and Sun, Simeng and Ginsburg, Boris},
  journal={arXiv preprint arXiv:2410.01131},
  year={2024}
}

@article{yuan2025native,
  title={Native sparse attention: Hardware-aligned and natively trainable sparse attention},
  author={Yuan, Jingyang and Gao, Huazuo and Dai, Damai and Luo, Junyu and Zhao, Liang and Zhang, Zhengyan and Xie, Zhenda and Wei, YX and Wang, Lean and Xiao, Zhiping and others},
  journal={arXiv preprint arXiv:2502.11089},
  year={2025}
}

@article{Zhu2024HyperConnections,
    title   = {Hyper-Connections},
    author  = {Defa Zhu and Hongzhi Huang and Zihao Huang and Yutao Zeng and Yunyao Mao and Banggu Wu and Qiyang Min and Xun Zhou},
    journal = {ArXiv},
    year    = {2024},
    volume  = {abs/2409.19606},
    url     = {https://api.semanticscholar.org/CorpusID:272987528}
}

@misc{puigcerver2023sparse,
    title 	= {From Sparse to Soft Mixtures of Experts}, 
    author 	= {Joan Puigcerver and Carlos Riquelme and Basil Mustafa and Neil Houlsby},
    year 	= {2023},
    eprint 	= {2308.00951},
    archivePrefix = {arXiv},
    primaryClass = {cs.LG}
}

@article{elhage2021mathematical,
   title={A Mathematical Framework for Transformer Circuits},
   author={Elhage, Nelson and Nanda, Neel and Olsson, Catherine and Henighan, Tom and Joseph, Nicholas and Mann, Ben and Askell, Amanda and Bai, Yuntao and Chen, Anna and Conerly, Tom and DasSarma, Nova and Drain, Dawn and Ganguli, Deep and Hatfield-Dodds, Zac and Hernandez, Danny and Jones, Andy and Kernion, Jackson and Lovitt, Liane and Ndousse, Kamal and Amodei, Dario and Brown, Tom and Clark, Jack and Kaplan, Jared and McCandlish, Sam and Olah, Chris},
   year={2021},
   journal={Transformer Circuits Thread},
   note={https://transformer-circuits.pub/2021/framework/index.html}
}

@inproceedings{hatamizadeh2022unetr,
  title={Unetr: Transformers for 3d medical image segmentation},
  author={Hatamizadeh, Ali and Tang, Yucheng and Nath, Vishwesh and Yang, Dong and Myronenko, Andriy and Landman, Bennett and Roth, Holger R and Xu, Daguang},
  booktitle={Proceedings of the IEEE/CVF winter conference on applications of computer vision},
  pages={574--584},
  year={2022}
}

@article{jordan1994hierarchical,
  title={Hierarchical mixtures of experts and the EM algorithm},
  author={Jordan, Michael I and Jacobs, Robert A},
  journal={Neural computation},
  volume={6},
  number={2},
  pages={181--214},
  year={1994},
  publisher={MIT Press}
}

@inproceedings{ainslie-etal-2023-gqa,
    title = "{GQA}: Training Generalized Multi-Query Transformer Models from Multi-Head Checkpoints",
    author = "Ainslie, Joshua  and
      Lee-Thorp, James  and
      de Jong, Michiel  and
      Zemlyanskiy, Yury  and
      Lebron, Federico  and
      Sanghai, Sumit",
    editor = "Bouamor, Houda  and
      Pino, Juan  and
      Bali, Kalika",
    booktitle = "Proceedings of the 2023 Conference on Empirical Methods in Natural Language Processing",
    month = dec,
    year = "2023",
    address = "Singapore",
    publisher = "Association for Computational Linguistics",
    url = "https://aclanthology.org/2023.emnlp-main.298/",
    doi = "10.18653/v1/2023.emnlp-main.298",
    pages = "4895--4901"
}

@inproceedings{liu2021swin,
  title={Swin transformer: Hierarchical vision transformer using shifted windows},
  author={Liu, Ze and Lin, Yutong and Cao, Yue and Hu, Han and Wei, Yixuan and Zhang, Zheng and Lin, Stephen and Guo, Baining},
  booktitle={Proceedings of the IEEE/CVF international conference on computer vision},
  pages={10012--10022},
  year={2021}
}

@inproceedings{shoemake1985animating,
  title={Animating rotation with quaternion curves},
  author={Shoemake, Ken},
  booktitle={Proceedings of the 12th annual conference on Computer graphics and interactive techniques},
  pages={245--254},
  year={1985}
}

@article{stringer2021cellpose,
  title={Cellpose: a generalist algorithm for cellular segmentation},
  author={Stringer, Carsen and Wang, Tim and Michaelos, Michalis and Pachitariu, Marius},
  journal={Nature methods},
  volume={18},
  number={1},
  pages={100--106},
  year={2021},
  publisher={Nature Publishing Group US New York}
}

@article{pachitariu2025cellpose,
  title={Cellpose-SAM: superhuman generalization for cellular segmentation},
  author={Pachitariu, Marius and Rariden, Michael and Stringer, Carsen},
  journal={bioRxiv},
  pages={2025--04},
  year={2025},
  publisher={Cold Spring Harbor Laboratory}
}

@article{dosovitskiy2020image,
  title={An image is worth 16x16 words: Transformers for image recognition at scale},
  author={Dosovitskiy, Alexey and Beyer, Lucas and Kolesnikov, Alexander and Weissenborn, Dirk and Zhai, Xiaohua and Unterthiner, Thomas and Dehghani, Mostafa and Minderer, Matthias and Heigold, Georg and Gelly, Sylvain and others},
  journal={arXiv preprint arXiv:2010.11929},
  year={2020}
}

@inproceedings{zhai2022scaling,
  title={Scaling vision transformers},
  author={Zhai, Xiaohua and Kolesnikov, Alexander and Houlsby, Neil and Beyer, Lucas},
  booktitle={Proceedings of the IEEE/CVF conference on computer vision and pattern recognition},
  pages={12104--12113},
  year={2022}
}

@inproceedings{dehghani2023scaling,
  title={Scaling vision transformers to 22 billion parameters},
  author={Dehghani, Mostafa and Djolonga, Josip and Mustafa, Basil and Padlewski, Piotr and Heek, Jonathan and Gilmer, Justin and Steiner, Andreas Peter and Caron, Mathilde and Geirhos, Robert and Alabdulmohsin, Ibrahim and others},
  booktitle={International conference on machine learning},
  pages={7480--7512},
  year={2023},
  organization={PMLR}
}

@inproceedings{kraus2024masked,
  title={Masked autoencoders for microscopy are scalable learners of cellular biology},
  author={Kraus, Oren and Kenyon-Dean, Kian and Saberian, Saber and Fallah, Maryam and McLean, Peter and Leung, Jess and Sharma, Vasudev and Khan, Ayla and Balakrishnan, Jia and Celik, Safiye and others},
  booktitle={Proceedings of the IEEE/CVF Conference on Computer Vision and Pattern Recognition},
  pages={11757--11768},
  year={2024}
}

@inproceedings{tongzhouw2020hypersphere,
  title={Understanding Contrastive Representation Learning through Alignment and Uniformity on the Hypersphere},
  author={Wang, Tongzhou and Isola, Phillip},
  booktitle={International Conference on Machine Learning},
  organization={PMLR},
  pages={9929--9939},
  year={2020}
}

@inproceedings{he2016deep,
  title={Deep residual learning for image recognition},
  author={He, Kaiming and Zhang, Xiangyu and Ren, Shaoqing and Sun, Jian},
  booktitle={Proceedings of the IEEE conference on computer vision and pattern recognition},
  pages={770--778},
  year={2016}
}

@article{christiansen2018silico,
  title={In silico labeling: predicting fluorescent labels in unlabeled images},
  author={Christiansen, Eric M and Yang, Samuel J and Ando, D Michael and Javaherian, Ashkan and Skibinski, Gaia and Lipnick, Scott and Mount, Elliot and O’neil, Alison and Shah, Kevan and Lee, Alicia K and others},
  journal={Cell},
  volume={173},
  number={3},
  pages={792--803},
  year={2018},
  publisher={Elsevier}
}

@article{ounkomol2018label,
  title={Label-free prediction of three-dimensional fluorescence images from transmitted-light microscopy},
  author={Ounkomol, Chawin and Seshamani, Sharmishtaa and Maleckar, Mary M and Collman, Forrest and Johnson, Gregory R},
  journal={Nature methods},
  volume={15},
  number={11},
  pages={917--920},
  year={2018},
  publisher={Nature Publishing Group US New York}
}

@article{rivenson2019phasestain,
  title={PhaseStain: the digital staining of label-free quantitative phase microscopy images using deep learning},
  author={Rivenson, Yair and Liu, Tairan and Wei, Zhensong and Zhang, Yibo and De Haan, Kevin and Ozcan, Aydogan},
  journal={Light: Science \& Applications},
  volume={8},
  number={1},
  pages={23},
  year={2019},
  publisher={Nature Publishing Group UK London}
}

@inproceedings{cciccek20163d,
  title={3D U-Net: learning dense volumetric segmentation from sparse annotation},
  author={{\c{C}}i{\c{c}}ek, {\"O}zg{\"u}n and Abdulkadir, Ahmed and Lienkamp, Soeren S and Brox, Thomas and Ronneberger, Olaf},
  booktitle={International conference on medical image computing and computer-assisted intervention},
  pages={424--432},
  year={2016},
  organization={Springer}
}

@inproceedings{milletari2016v,
  title={V-net: Fully convolutional neural networks for volumetric medical image segmentation},
  author={Milletari, Fausto and Navab, Nassir and Ahmadi, Seyed-Ahmad},
  booktitle={2016 fourth international conference on 3D vision (3DV)},
  pages={565--571},
  year={2016},
  organization={Ieee}
}

@article{Isensee_2020,
   title={nnU-Net: a self-configuring method for deep learning-based biomedical image segmentation},
   volume={18},
   ISSN={1548-7105},
   url={http://dx.doi.org/10.1038/s41592-020-01008-z},
   DOI={10.1038/s41592-020-01008-z},
   number={2},
   journal={Nature Methods},
   publisher={Springer Science and Business Media LLC},
   author={Isensee, Fabian and Jaeger, Paul F. and Kohl, Simon A. A. and Petersen, Jens and Maier-Hein, Klaus H.},
   year={2020},
   month=dec, pages={203–211}
}

@article{chen2021transunet,
  title={Transunet: Transformers make strong encoders for medical image segmentation},
  author={Chen, Jieneng and Lu, Yongyi and Yu, Qihang and Luo, Xiangde and Adeli, Ehsan and Wang, Yan and Lu, Le and Yuille, Alan L and Zhou, Yuyin},
  journal={arXiv preprint arXiv:2102.04306},
  year={2021}
}

@inproceedings{hatamizadeh2021swin,
  title={Swin unetr: Swin transformers for semantic segmentation of brain tumors in mri images},
  author={Hatamizadeh, Ali and Nath, Vishwesh and Tang, Yucheng and Yang, Dong and Roth, Holger R and Xu, Daguang},
  booktitle={International MICCAI brainlesion workshop},
  pages={272--284},
  year={2021},
  organization={Springer}
}

@article{beltagy2020longformer,
  title={Longformer: The long-document transformer},
  author={Beltagy, Iz and Peters, Matthew E and Cohan, Arman},
  journal={arXiv preprint arXiv:2004.05150},
  year={2020}
}

@article{zhu2020deformable,
  title={Deformable detr: Deformable transformers for end-to-end object detection},
  author={Zhu, Xizhou and Su, Weijie and Lu, Lewei and Li, Bin and Wang, Xiaogang and Dai, Jifeng},
  journal={arXiv preprint arXiv:2010.04159},
  year={2020}
}

@article{zhang2019root,
  title={Root mean square layer normalization},
  author={Zhang, Biao and Sennrich, Rico},
  journal={Advances in neural information processing systems},
  volume={32},
  year={2019}
}

@article{zaheer2020big,
  title={Big bird: Transformers for longer sequences},
  author={Zaheer, Manzil and Guruganesh, Guru and Dubey, Kumar Avinava and Ainslie, Joshua and Alberti, Chris and Ontanon, Santiago and Pham, Philip and Ravula, Anirudh and Wang, Qifan and Yang, Li and others},
  journal={Advances in neural information processing systems},
  volume={33},
  pages={17283--17297},
  year={2020}
}

@inproceedings{zhou2018unet++,
  title={Unet++: A nested u-net architecture for medical image segmentation},
  author={Zhou, Zongwei and Rahman Siddiquee, Md Mahfuzur and Tajbakhsh, Nima and Liang, Jianming},
  booktitle={International workshop on deep learning in medical image analysis},
  pages={3--11},
  year={2018},
  organization={Springer}
}

@article{lepikhin2020gshard,
  title={Gshard: Scaling giant models with conditional computation and automatic sharding},
  author={Lepikhin, Dmitry and Lee, HyoukJoong and Xu, Yuanzhong and Chen, Dehao and Firat, Orhan and Huang, Yanping and Krikun, Maxim and Shazeer, Noam and Chen, Zhifeng},
  journal={arXiv preprint arXiv:2006.16668},
  year={2020}
}

@article{fedus2022switch,
  title={Switch transformers: Scaling to trillion parameter models with simple and efficient sparsity},
  author={Fedus, William and Zoph, Barret and Shazeer, Noam},
  journal={Journal of Machine Learning Research},
  volume={23},
  number={120},
  pages={1--39},
  year={2022}
}

@article {Chen491035,
	author = {Chen, Jianxu and Ding, Liya and Viana, Matheus P. and Lee, HyeonWoo and Sluezwski, M. Filip and Morris, Benjamin and Hendershott, Melissa C. and Yang, Ruian and Mueller, Irina A. and Rafelski, Susanne M.},
	title = {The Allen Cell and Structure Segmenter: a new open source toolkit for segmenting 3D intracellular structures in fluorescence microscopy images},
	elocation-id = {491035},
	year = {2020},
	doi = {10.1101/491035},
	publisher = {Cold Spring Harbor Laboratory},
	URL = {https://www.biorxiv.org/content/early/2020/12/13/491035},
	eprint = {https://www.biorxiv.org/content/early/2020/12/13/491035.full.pdf},
	journal = {bioRxiv}
}
}
\end{document}